\documentclass[a4paper,english,11pt]{article}

\usepackage{graphicx}
\usepackage{xspace}
\usepackage{mathtools}
\usepackage{authblk}
\usepackage{subcaption}
\usepackage{todonotes}
\usepackage{amsmath}
\usepackage{units}
\usepackage[round]{natbib}
\usepackage{hyperref}
\hypersetup{
    colorlinks=true,       
    linkcolor=black,          
    citecolor=black,        
    filecolor=black,      
    urlcolor=blue           
}
\newcommand{\AAn}{\ensuremath{\text{AA}\xspace}}
\newcommand{\ABn}{\ensuremath{\text{AB}\xspace}}
\newcommand{\BAn}{\ensuremath{\text{BA}\xspace}}
\newcommand{\BBn}{\ensuremath{\text{BB}\xspace}}
\newcommand{\OO}{\ensuremath{\text{OO}\xspace}}
\newcommand{\OX}{\ensuremath{\text{OX}\xspace}}
\newcommand{\XO}{\ensuremath{\text{XO}\xspace}}
\newcommand{\XX}{\ensuremath{\text{XX}\xspace}}

\AtBeginDocument{}

\author{%
  Roman Miletitch,$^1$ Andreagiovanni Reina,$^{1,2}$ Marco Dorigo$^1$ and  Vito Trianni$^3$\\
  $^1$IRIDIA, Universit\'{e} Libre de Bruxelles, Brussels, Belgium\\
  $^2$Department of Computer Science, University of Sheffield, Sheffield, UK\\
  $^3$ISTC, Italian National Research Council, Rome, Italy
}

\begin{document}

\title{Emergent naming conventions in a foraging robot swarm}

\maketitle
\begin{abstract}
In this study, we investigate the emergence of naming conventions
within a swarm of robots that collectively forage, that is, collect
resources from multiple sources in the environment. While foraging,
the swarm explores the environment and makes a collective decision on
how to exploit the available resources, either by selecting a single
source or concurrently exploiting more than one. At the same time, the
robots locally exchange messages in order to agree on how to name each
source. Here, we study the correlation between the task-induced
interaction network and the emergent naming conventions. In
particular, our goal is to determine whether the dynamics of the
interaction network are sufficient to determine an emergent vocabulary
that is potentially useful to the robot swarm. To be useful,
linguistic conventions need to be compact and meaningful, that
is, to be the minimal description of the relevant features of the
environment and of the made collective decision. We show that, in
order to obtain a useful vocabulary, the task-dependent interaction
network alone is not sufficient but it must be combined with a 
correlation between language and foraging dynamics. On the basis of
these results, we propose a decentralised algorithm for collective
categorisation which enables the swarm to achieve a useful---compact
and meaningful---naming of all the available sources. Understanding
how useful linguistic conventions emerge contributes to the design of
robot swarms with potentially improved autonomy, flexibility, and
self-awareness.

\end{abstract}

\section{Introduction}

The development of advanced forms of communication---i.e., a primitive
form of language---can help robots in a swarm to share relevant
information about the task execution, adapting it to the current
activities and environmental contingencies experienced by the robots
\citep{CambierMiletitch2019}. Indeed, linguistic conventions can be
useful to describe the environment and the task execution progress in
a compact way, supporting the coordination within the swarm. Among the
tasks relevant for swarm robotics, foraging---a task often observed in
natural self-organising systems \citep{bailis2010positional,
  saleh2007traplining}---is certainly one among the most studied
\citep{Ducatelle:2014hv,Ferrante:2015kq,miletitch2018balancing,talamali2020}, as it lends itself to represent multiple realistic applications like
mining, search-and-rescue or logistics.  While foraging, the swarm
needs to explore an environment and decide which source to exploit
among several available. In such context, linguistic
conventions can provide compact ways of uniquely identifying relevant
aspects of the environment (e.g., different terms to identify
different sources from which to forage), which can evolve to adapt to
a changing landscape (e.g., assigning new terms to newly discovered
sources, or dropping terms associated with depleted sources), hence
maximising the communication efficiency. Moreover, an evolving
language can contain sequences of terms, providing
swarms the ability to decide on the most useful course of action
(e.g., a sequence of sources from which to forage).

To make language evolution possible, however, robots in a swarm need
to interact and agree on the terms to be used and their meaning. This
is the realm of \emph{language games}, that is, computational models
developed to understand the emergence of language through
communication and self-organisation \citep{steels2001language,
  baronchelli2010modeling, Spranger:2013iq}. As in swarm robotics communication is often local and intermittent, complex and dynamical interaction networks among robots emerge. A language
game played in these conditions would have its dynamics largely
affected by the network topology resulting from the task execution
\citep{Loreto:2011jn}.  In this paper, we study the correlation between the task-induced interaction network and the evolving language. Indeed, the outcome of the
language game can be correlated with both the intrinsic dynamics and
outcome of the task itself, and the features of the environment in
which the task is carried out.  When such correlations are present,
the linguistic conventions resulting from the language game are
semantically grounded onto the task and its environment, and can
therefore be exploited for the accomplishment of the task itself. Some
experiments have explored semantic connections between language games
and the physical spaces in which they are played
\citep{steels1995self, Spranger:2013iq}. However, applications in swarm robotics are
still limited \citep{CambierMiletitch2019}, and only a few experiments
with a self-organised aggregation problem can be reported to date
\citep{cambier2018embodied}.  

In this paper, we demonstrate how language games can be grounded onto
the execution of a foraging task. Specifically, we show that the
task-induced interaction network is not sufficient \emph{per se} in
determining the conditions for semantically grounding the emergent
linguistic conventions onto the task. However, we show that such
grounding is possible when the language game is played by robots
actually exploiting a source. The understanding of the language
dynamics leads us to define a category game tailored to better
represent the different sources distributed in space, as long as these
are relevant to the foraging task.

The paper is organised as follows. In Section~\ref{sec:mng-swarm}, we
discuss how language games can be meaningfully played by a robot swarm
engaged in a source exploitation task. In
Section~\ref{sec:experimental-setup}, we present the experimental
setup. In Section~\ref{sec:match-compl-vocab}, we show how the
dynamics of the interaction network can lead to emergent linguistic
conventions. Then, in Section~\ref{sec:study-swarms-spatial}, we
analyse the properties of the interaction network, suggesting that it
meaningfully supports the evolution of useful linguistic
conventions. Finally, in Section~\ref{sec:from-vocab-dict} we present
the category game introduced to better support self-organised
foraging. Finally, Section~\ref{sec:conclusion} concludes the
paper.


\section{Language games in foraging robot swarm}
\label{sec:mng-swarm}


In swarm robotics, coordination and self-organisation allow groups of
robots to be more efficient than isolated robots in performing a given
task~\citep{DorBirBra2014:sch-sr,DorigoEtAl:SR2020}. The collaborative
processes designed for robot swarms are often inspired by social
insects and other group-living animals~\citep{Brambilla:2013ja,Trianni:2015if}.  Communication is one fundamental aspect for self-organisation, and  can be either indirect
(e.g., stigmergy) or direct. Both types of communication are
encountered in animal societies, such as the pheromone trails used by
ants \citep{beekman2001phase} or the waggle dance used by honey bees
\citep{biesmeijer2001explo}. These communication mechanisms have been
implemented with success in swarm robotics systems, for example using
indirect stigmergic interactions
\citep{Holland:1999uc,beckers2000fom,AllBhaElf-etal2014:ants},
pheromones \citep{fujisawa2014designing, talamali2020} and direct
communication \citep[][]{Gutierrez:2010ea,
  miletitch2018balancing}. While efficient, these communication
mechanisms are usually designed for a specific task/environment
\citep[e.g., application in warehouses, see][]{stiefel2004dist} and convey specific pieces of information, hence
limiting the system flexibility.

Researchers aimed to add more plasticity to the communication process,
for instance by exploiting an evolutionary process to design at the
same time signals and adapted responses \citep{marocco2007emergence,
  floreano2007evolutionary}. The resulting communication mechanisms
are very well adapted to the tasks and environmental conditions
encountered during training, and also show some generalisation
abilities. However, the characteristics of the obtained communication
mechanisms remain very simple, with few signals and responses to
signals that cannot easily scale up to more complex environments
and/or tasks. A possibility to provide more complex communication
abilities to a robotic system comes from models of natural language
evolution \citep[][]{wang2005invasion, sole2010diversity}.
%

A popular approach to the study of language dynamics is represented by
language games played by a population of 
agents/robots, with the purpose of mimicking real-world linguistic
interactions leading to the emergence of a structured language.
Various kinds of language games have been proposed to date, from
imitation games \citep{billard1997learning} to guessing games
\citep{steels2001language} and category games
\citep{puglisi2008cultural, baronchelli2010modeling}.
%
One game in particular has received a lot of attention: the
\emph{naming game} \citep{steels1995self, Steels:2003wj}. In this
game, two or more robots interact to assign a unique name to a set of
objects. At each interaction, one robot is chosen as a speaker and
another as a listener. The speaker chooses a referring object and an
associated word from its vocabulary---or invents one when no word is
available---and then transmits it to the listener. If the listener
knows the word, then the game is a success, and both agents remove all
other words associated to the chosen object from their vocabulary,
keeping only the shared word. If instead the listener does not know
the received word, then the game fails, and the listener adds this new
word to its vocabulary. We use in our study a specific version of this
game: the minimal naming game \citep[MNG,
see][]{baronchelli2006sharp}. Here, focus is given only to reaching
consensus on a single world within a population of communicating
agents. Specifically, we consider an implementation in which the
speaker broadcasts its word to all agents in his neighbourhood, while
the listener is the only agent that updates the vocabulary upon
success or failure of a game \citep{baronchelli2011role}.

As naming games are based on interactions between pairs of speaker and 
listener agents, the time to achieve consensus and the underlying dynamics are directly
linked to the topology of the interaction network. In non-embodied
implementations, the link between topology and language dynamics have
been extensively studied \citep[e.g., fully-connected regular,
small-world or random geometric networks,
see][]{baronchelli2007role,lu2008naming}. Embodied
implementations can be divided in two cases. On the one hand, a
population of virtual agents can use a small number of robots
\citep[sometimes reduced to two, as
in][]{Spranger:2013iq} to play the naming game, so
that at each iteration, agents are selected and assigned to robots in
order to record physical interactions among them. On the other hand,
the naming game can be played among a population of embodied mobile
agents \citep[][]{Baronchelli:2012dq, trianni2016emergence} that
interact locally with each other according to a topology of
interactions that is the direct result of the mobility pattern of the
agents induced from the task being executed.

In this study, the MNG is played on top of a self-organised foraging
task. When foraging, a swarm needs to explore the environment,
identify and evaluate the available sources and make decisions on
which source to exploit, going through different transitory states
before reaching an equilibrium 
\citep[e.g., convergence on one single source to exploit or split/load-balance among many, as in][]{miletitch2018balancing}. Similar behaviours provide a complex
and time varying interaction network among robots, which can be
exploited to support linguistic interactions among agents. Our main
goal is to study whether the dynamics of the interaction network are
sufficient to determine language dynamics that represent features of
the task execution (e.g., choice of one or the other source), of the
environment (e.g., the presence of more than one sources, each
associated to a different word), or both.
To this end,
we run experiments with two versions of the MNG. Beside the classic
MNG, we play a version where the creation of words is linked with the
discovery of sources by exploring robots. In this setup, we study
how well the robots manage to have an accurate description of their
surroundings, that is both complete (a word for each source) and
correct (no misnomer) for as long as each source is relevant to the
swarm, where relevance is measured as the number of robots actively
foraging from the source (see Section~\ref{sec:experimental-setup}).
Our goal is to understand how the swarm interaction topology influences 
the language dynamics, and how the creation of words is correlated 
with the robots foraging from a source.


\section{Experimental setup}
\label{sec:experimental-setup}

\begin{figure}[!b]
    \centering
    \includegraphics[width=0.9\textwidth]{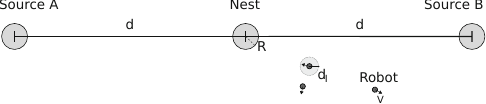}
    \caption{Graphical representation of the environment. sources A
      and B are each located at the same distance $d = \unit[2.5]{m}$
      from the nest. All the three areas have radius
      $R = \unit[0.3]{m}$. Robots move at constant speed
      $v = \unit[0.1]{ms^{-1}}$ and can communicate with neighbours
      within a range $d_I = \unit[0.2]{m}$.}
    \label{fig:arena}
\end{figure}

In this study, the goal of the swarm is to play a MNG
while identifying and exploiting either of two sources (referred to
as source A and source B) placed at the opposite side of a home
area (referred to as nest, see Figure~\ref{fig:arena}). The
environment is a 2D infinite plane without obstacles, and both nest and sources
have circular shape with radius $R = \unit[0.3]{m}$. Each source is
located at the same distance $d = \unit[2.5]{m}$ from the nest.

\subsection{ Robots and simulations }
\label{sec:robots-simulations}
Experiments are run in simulation using ARGoS
\citep{Pinciroli:2012dc}. In our study, we use this simulator to model
a swarm of 50 e-puck robots \citep{Mondada:2009tw}. E-pucks have a differential drive motion with a maximum linear speed of $v = \unitfrac[0.1]{m}{s}$, and the wheels' rotation is measured by an encoder.
Avoidance of other robots is done at short range ($\approx \unit[10]{cm}$) using infrared proximity sensors and at longer range ($\approx \unit[1]{m}$) using the infrared range and bearing system \citep{GutierrezRAB}. 
The obstacle avoidance behaviour has been optimised to minimise the effects of robot density and congestion and to support the ability to navigate back and forth between sources, as detailed in a previous study
\citep{Miletitch:2013bk}. 
Robots perceive nest and sources only when they are located in the corresponding areas by means of infrared ground sensors, that robots use to differentiate between the white colour of the
floor, the grey colour of the sources and the black colour of the nest. We assume here that robots start from the nest without any knowledge about sources, which need to be located through exploration.
Robots can locally broadcast short messages through the infrared range and bearing system within a range that is limited to $d_I = \unit[0.2]{m}$ (indicated by the dotted circle around the robot in Figure
\ref{fig:arena}). 
Robots can broadcast a message at regular intervals of \unit[0.1]{s} with no re-broadcast of information received (no multi-hop communication). They keep track of the position of nest and known sources through odometry. The error on positioning produced through this tracking method can be efficiently compensated through social odometry \citep{Gutierrez:2010ea,Miletitch:2013bk}. Owing to this, in this study we neglect odometry errors and focus on the interplay between motion and language dynamics.

At the beginning of the experiment, robots are uniformly distributed within a \unit[0.8]{m} side square centered on the nest. During the first \unit[200]{s}, robots perform a blind random walk during which
they do not communicate or search for sources. This allows us to
neglect the initial transitory phase in which robots are too densely distributed around the nest, allowing us to study the system dynamics after the robots spread out in the environment according to their search pattern. This assures that---whatever the experimental condition---the initial distribution of robots does not severely impact the final outcome. In the following experiments, unless mentioned otherwise, we perform $100$ runs for each experimental setup. These runs last until language convergence, which, depending on internal parameters, can take up to $\unit[12000]{s}$.

\subsection{Individual and collective behaviour}
\label{sec:indiv-coll-behav}

\subsubsection{Source exploitation}

The desired swarm behaviour (localization and exploitation of
sources) takes inspiration from the decision-making process
displayed by house-hunting honeybees---also know as nest-site
selection \citep[NSS, see][]{Pais:2013ek,Seeley108,Reina:PRE:2017}. The spatial
dynamics during foraging resulting from the NSS process have been
studied by \citet{Reina:2015gs} and \citet{miletitch2018balancing}. Here, we make use of the
individual robot behaviour from the former \citep{Reina:2015gs}, which was
designed for the e-puck robots following a design pattern based on the
NSS process \citep{Reina:2015hu}. According to this design pattern, a robot
is considered to be committed to a source when it knows its
location, and hence moves back and forth between the source and the
nest. Otherwise, a robot is considered uncommitted and explores the
arena searching for a source. Robots committed to source A (B) are
considered to belong to the population $\mathcal{P}_A$
($\mathcal{P}_B$), while uncommitted robots belong to the population
$\mathcal{P}_U$, all summing up to $N$ robots: 
$|\mathcal{P}_A| + |\mathcal{P}_B| + |\mathcal{P}_U| = N$.

Four concurrent processes determine the individual behaviour, two for transitions between uncommitted and committed states, and two for the opposite.  
An uncommitted robot turns committed either through \textbf{discovery} or through \textbf{recruitment}. The former takes place when the robot enters the area of a source. The latter takes place with probability $P_\rho$ when a robot receives the information about a source known by a committed neighbour. 
Conversely, a committed robot turns uncommitted either through \textbf{abandonment} or through \textbf{cross-inhibition}. The former takes place anytime with a fixed probability $P_\alpha$ per time-step. The latter takes place with probability $P_\sigma$ upon iteraction with a neighbouring robot committed to a different source. Cross-inhibition introduces a negative feedback loop that helps the system break the symmetry and leads to a choice between two identical sources \citep[see][for more details]{Reina:2015gs,Reina:2015hu}. 
In our study, recruitment and cross-inhibition happen only upon communication with other robots when located into the nest. Differently from \citet{Reina:2015gs}, we set the probability of abandonment $P_\alpha$ to zero, so that the only way for robots to become uncommitted is through cross-inhibition. This favours the attainment of a consensus state in which all robots within the swarm are committed to the one or the other source \citep{Reina:2015hu}.

The actual movements of the robot are governed by the following basic
behaviours. When uncommitted, the robots explore the arena, performing
a correlated random walk \citep{Dimidov:2016gp}, and have a fixed and
small probability at every control step to return to the nest. When
committed, the robots enter an exploitation loop where they move back
and forth between the known source and the nest \citep[see][for a
detailed description]{Reina:2015gs}.


\begin{figure}[!t]
    \centering
    \includegraphics[width=0.98\textwidth]{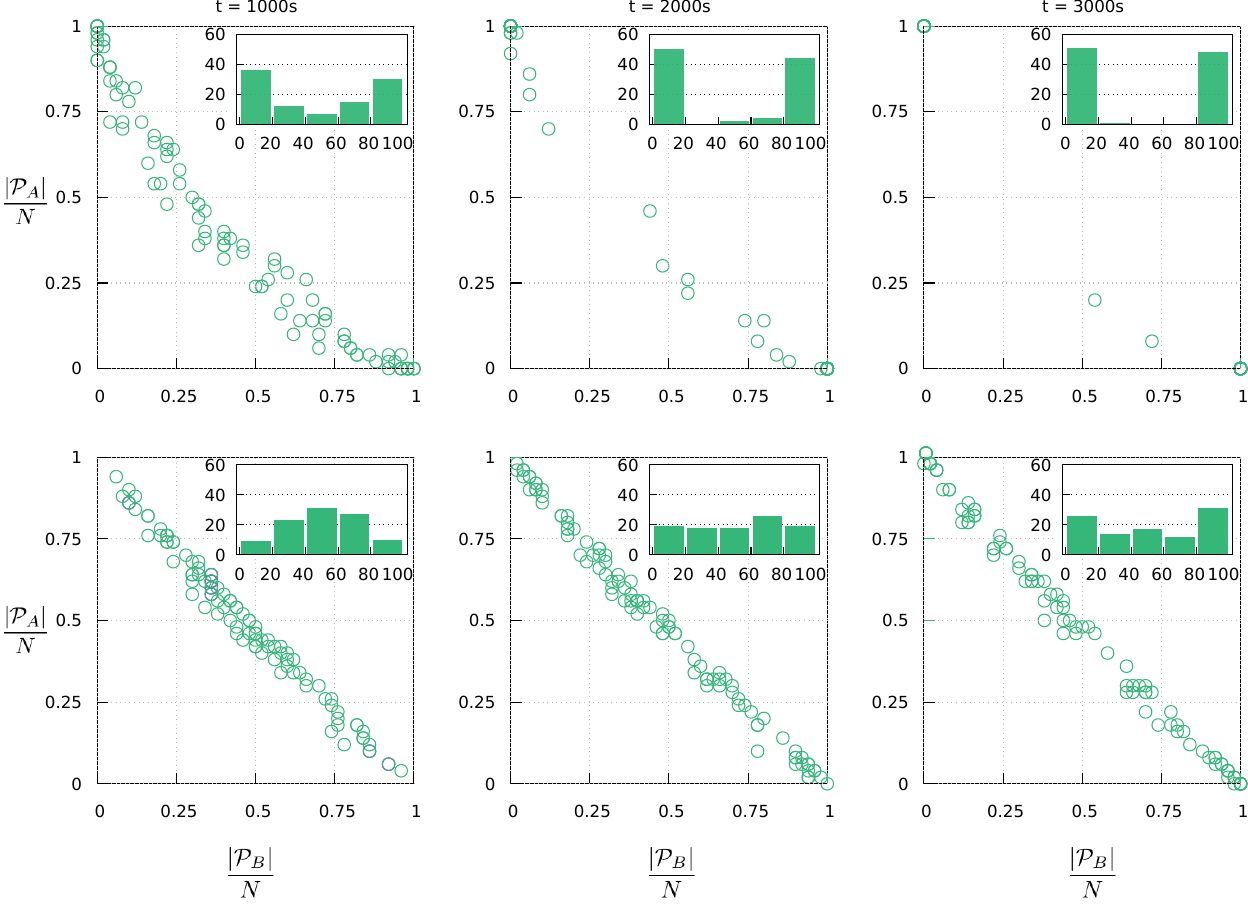}
    \caption{Distribution of robots in a swarm as a percentage of
      robots committed to source A (y axis) and B (x axis) for 100
      independent runs. Each column displays the distribution at
      different time steps. The insets show the histogram of the
      frequencies of runs with respect to the percentage of robots
      committed to A. Top row: \textbf{strong cross-inhibition} with
      $P_\rho = 0.7$ and $P_\sigma = 0.7$, robots can change
      commitment and eventually the swarm converges toward either
      source A or B. Bottom row: \textbf{weak cross-inhibition} with
      $P_\rho = 0.7$ and $P_\sigma = 0.1$, the dynamic is much
      slower. Over the duration of our experiments, each run ends up
      with a different distribution of robots among sources, with
      points close to the diagonal representing low number of uncommitted robots.}
    \label{fig:nss_end}
\end{figure}

Depending on the value of $P_\rho$ and $P_\sigma$, the swarm displays
different dynamics and different final distributions of robots among
the populations $\mathcal{P}_U$, $\mathcal{P}_A$ and $\mathcal{P}_B$.
In this study, we focus on two specific cases: \textbf{strong
  cross-inhibition} and \textbf{weak cross-inhibition}. In the strong
case ($P_\sigma = 0.7$, Figure~\ref{fig:nss_end} top row) the swarm
rapidly converges to a consensus for the one or the other source, whereas the weak
case ($P_\sigma = 0.1$, see Figure~\ref{fig:nss_end} bottom row) leads
to slower dynamics \citep{Reina:DARS:2016}. Given enough time the swarm would end up
converging to a consensus for a single source. However, over the duration of our experiments, the
swarm did not break the symmetry but splits between the two sources (see
Figure~\ref{fig:nss_end}, bottom row). At any time, with or without
consensus, we define the source with the highest number of
committed robots (relative majority) as the ``selected'' source. We define $O\in\{A,B\}$
as the selected source and $X\in\{A,B\}$ as the non-selected
source, and $\mathcal{P}_O$ and $\mathcal{P}_X$ as the respective
populations, with $\mathcal{P}_O\geq \mathcal{P}_X$.

\subsubsection{Minimal naming game}

The language game played by the robots in our study is an
implementation of the minimal naming game (MNG) for mobile
agents/robots
\citep{baronchelli2006sharp,Baronchelli:2012dq,trianni2016emergence}. Each
robot starts with an empty inventory. At each time step  (of length $\tau_c=\unit[100]{ms}$), each robot
has a probability $P_s$ of becoming a speaker (here,
$P_s \in \{0.0003, 0.0006, 0.001, 0.002\}$). These values of $P_s$ were selected so that foraging dynamics and language dynamics would share comparable time scales.
The language game is played
as follows: the speaker robot selects a word from its inventory
and broadcasts it to its neighbours. At each time step, if a robot
receives at least one message, it becomes a hearer robot. The hearer
selects one (and only one) word at random among those received and
checks it against its own inventory. If the hearer finds the selected
word in its inventory, the hearer keeps only that word in the
inventory while deleting all the others. If instead the hearer does
not find the selected word in its inventory, it updates its inventory
by adding the word \citep[see][for more details]{trianni2016emergence}.

\begin{figure}[!t] 
    \centering
    \includegraphics[width=0.75\textwidth]{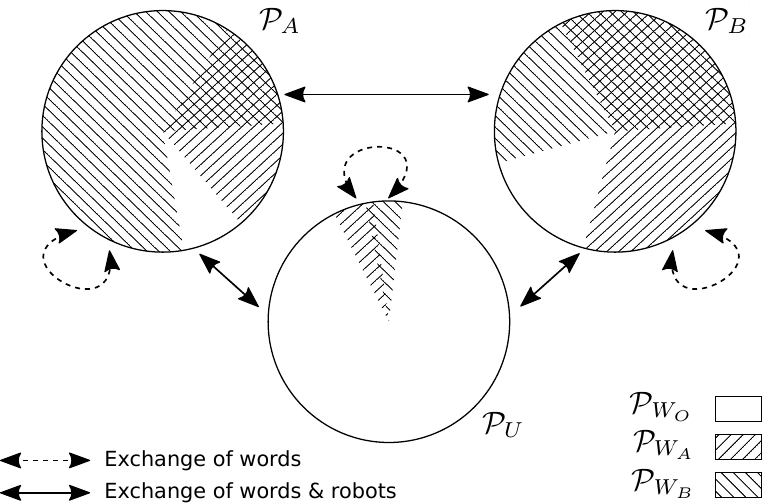}
    \caption{Diagram representing how the swarm can be split in
      different sub-populations with respect to the robots' commitment
      state and the word distribution. The circles represents the
      three populations with respect to the commitment state:
      ($\mathcal{P}_{U}$, $\mathcal{P}_{A}$ and
      $\mathcal{P}_{B}$). The fill patterns represent populations with
      respect to the robots' inventory ($\mathcal{P}_{W_O}$,
      $\mathcal{P}_{W_A}$ and $\mathcal{P}_{W_B}$). Note that, in
      general, $\mathcal{P}_{W_A}
      \cap\mathcal{P}_{W_B}\neq\emptyset$.
      Depending on the experimental setup, populations can exchange
      robots and words among themselves.}
    \label{fig:pop_AB}
\end{figure}

In this study, we consider two variants of the MNG, which differ in
the way in which words are generated. In one case (referred to as
\textbf{classic game}), the robots create a new word when becoming
speaker with an empty vocabulary. In the other (referred to as
\textbf{spatial game}), the robots create a new word when encountering a
source with an empty vocabulary.
In both cases, we associate each word with the closest source to the
robot at the time of the word creation, and we define $W_A$ ($W_B$)
the set of words associated with source $A$ ($B$). Note that, by
construction, $W_A \cap W_B = \emptyset$. Robots having in their
inventory any word $w \in W_A$ ($W_B$) constitute population
$\mathcal{P}_{W_A}$ ($\mathcal{P}_{W_B}$). Robots with no words
constitute population $\mathcal{P}_{W_O}$. In Figure~\ref{fig:pop_AB},
we depict a possible partition of robots among different
populations, both with respect to the commitment state and to their
vocabulary. Since a robot can have at a given time an inventory with
words originating in both source $A$ and $B$, the propriety
$\mathcal{P}_{W_A} \cap \mathcal{P}_{W_B} = \emptyset$ is not always
verified. Similarly, through exchanges of words and robots between the
different populations, at a given time the inventory of robots
committed to one source might contain a word associated with the
other source (resulting in $\mathcal{P}_A \neq \mathcal{P}_{W_A}$).
At any time, we can look at the population of robots that know words
associated with the source they are committed to, that is:
\begin{equation}
  \mathcal{P}_{M} = (\mathcal{P}_{W_A} \cap \mathcal{P}_{A}) \cup (\mathcal{P}_{W_B} \cap \mathcal{P}_{B}).
\end{equation}
Conversely, we can define the population of committed robots that know
words from a non-matching source: 
\begin{equation}
  \mathcal{P}_{S} = (\mathcal{P}_A \cap \mathcal{P}_{W_B}) \cup (\mathcal{P}_B \cap \mathcal{P}_{W_A}).
\end{equation}
Corresponding to the collectively selected source $O$ (see definition above), we
define the set of matching words $W_O$ and non-matching words $W_X$ as
follows:  
\begin{align}
    W_O&=\{w | (w \in W_A \wedge \mathcal{P}_A > \mathcal{P}_B) \vee ( w \in W_B \wedge \mathcal{P}_B > \mathcal{P}_A)\}\\
    W_X&=\{w | (w \in W_A \wedge \mathcal{P}_B > \mathcal{P}_A) \vee ( w \in W_B \wedge \mathcal{P}_A > \mathcal{P}_B)\}
\end{align}
We define:
\begin{itemize}
\item \textbf{polarisation}, the condition in which committed robots
  know only words associated with the source they are committed to,
  that is, when $\mathcal{P}_{S} = \emptyset$;
\item \textbf{vocabulary matching}, the condition in which only words
  associated with the selected source are retained within the swarm
  vocabulary, that is $W_X = \emptyset$ and $W_O\neq\emptyset$;
\item \textbf{vocabulary completeness}, the condition in which exactly
  one word associated with each source is retained within the swarm
  vocabulary, that is $|W_O| = 1$ and $|W_X| = 1$.
\end{itemize}

Given a sufficiently connected swarm, the MNG dynamics ensure that the
swarm will eventually converge to a final single-word vocabulary,
albeit after a very long time
\citep{baronchelli2006sharp,Baronchelli:2012dq,trianni2016emergence}.
According to the previous definitions, the final vocabulary can be
matching or not the selected source.

\section{Correctness and completeness of the swarm vocabulary}
\label{sec:match-compl-vocab}


In this section, we focus on the evolution of the swarm's vocabulary,
looking in particular to the provenance of the last words and their
relation to the selected source. As already discussed (see Figure
\ref{fig:nss_end}), the foraging dynamics lead to either the quick
selection of a single source, or to the swarm being split between the two sources, possibly for a long time. This means that, apart for a few cases and
random fluctuations, there will always be a source that is
selected---albeit temporarily---by the swarm. In certain settings, the
swarm may forage from both sources for a long time, hence vocabulary
completeness may be observed. In other cases, the swarm will quickly
converge to exploit a single source, and vocabulary matching is
expected. In any case, interactions between different populations of
robots are frequent, ensuring that the language dynamics always
converge to a single-word vocabulary.


Here, we first focus on the patterns
observed when the vocabulary converges to one or two words, to determine
if matching and completeness are achieved. First, we analyse the provenance of
the final word $w_f$ to determine if it matches the selected
source or not (i.e., $w_f\in W_O$). As the distribution of robots
among sub-populations may sometimes change even after convergence to a
single-word dictionary (e.g., if the language dynamics are much faster
than the source selection dynamics), the final selected source may
also change. Hence, we consider the source selected at the time of
convergence to the final word $w_f$, no matter what happens later to
the population distribution. Similarly, we consider also the
second-last word $w_e$, to determine whether it was also matching the
selected source or not at the time in which only two words remained
within the whole swarm. Given such definitions, every run can end up
in one of the following four possibilities: 
\begin{align}
  \OO &: w_f \in W_O \wedge w_e \in W_O\\
  \OX &: w_f \in W_O \wedge w_e \in W_X\\
  \XO &: w_f \in W_X \wedge w_e \in W_O\\
  \XX &: w_f \in W_X \wedge w_e \in W_X
\end{align}
In case $\OO$ or $\OX$ is observed, the swarm has identified a final
word that matches the currently-selected source, although in the
$\OX$ case the second-last word was associated with the non-selected
source. The $\XO$ case represents a missed opportunity of matching,
as a matching word was still existing in the vocabulary and could have
been chosen. The $\XX$ case instead suggests that the association of
words to source does not reflect the current state of the source
selection. Both middle cases ($\OX$ and $\XO$) indicate a complete
vocabulary up until convergence on one word.

\begin{figure}[!t]
    \centering
    \includegraphics[width=\textwidth]{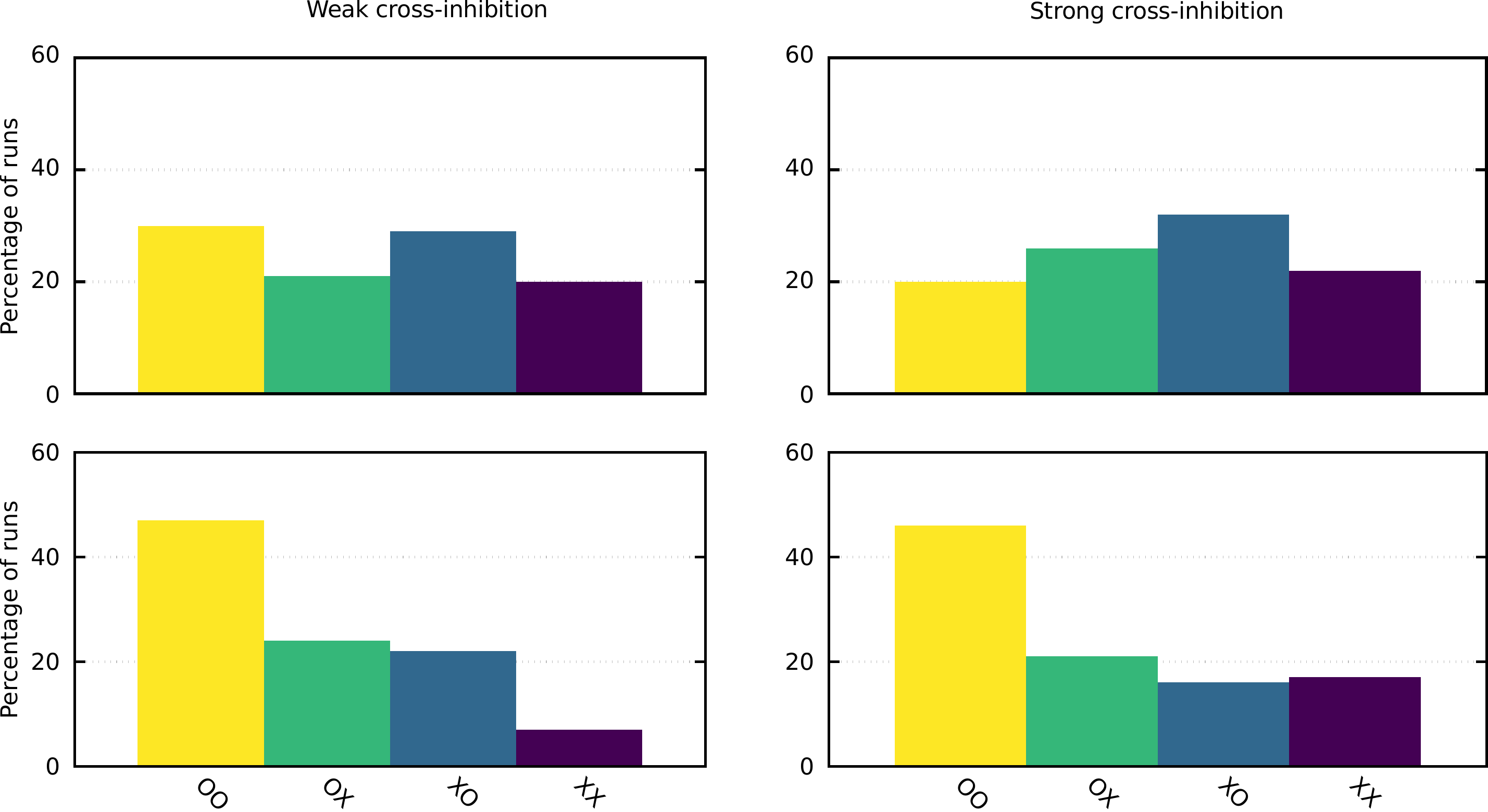}
    \caption{Empirical distribution over 100 runs of the occurrences of the last two words in the
      vocabulary within the four identified classes ($\OO$, $\OX$, $\XO$
      and $\XX$) representing words matching or not the selected
      source. The graph refers to the case with $P_s = 0.001$.  All
      other tested values of $P_s$ produce similar results (see
      supplementary Figure~S1). Top row: classic game. Bottom row: spatial game.}
    \label{fig:cv}
\end{figure}



Given these definitions, we study the influence of the language game
and the foraging dynamics over the provenance of the last two words of
the vocabulary. Figure~\ref{fig:cv} shows the frequency of each case
out of the 100 runs performed for each different experimental
condition.
When playing the classic game (top row in Figure~\ref{fig:cv}), the
swarm shows no tendency to favor a specific provenance for the final
two words, and a distribution close to uniform across the four
possible cases is observed. On the other hand, when playing the
spatial game (bottom row in Figure~\ref{fig:cv}), the swarm favours
words that match the selected source, both for the last and
second-last word. In particular, the $\OO$ state is strongly favoured
for both weak and strong cross-inhibition, and the $\XX$ state is
especially disfavoured when the weak cross-inhibition leads to slower
decision dynamics.
In conclusion, we clearly find that the spatial game, by making the creation
of words conditional to the discovery of sources,
determines a strong tendency to converge towards words that represent
the source that is ultimately selected. The naming process is
``correct'' as it best represents the source that is the most relevant
for the swarm. In about 40\% of the cases ($\OX + \XO$), the naming is
``complete'' as the last two words represent ``names'' for both the
available sources. This remains valid for different values of the
probability of speaking $P_s$, as shown in the supplementary
Figure~S1, suggesting that the spatial game is resilient to variations in the timescale of the language game.

\begin{figure}[t] 
    \centering
    \includegraphics[width=\textwidth]{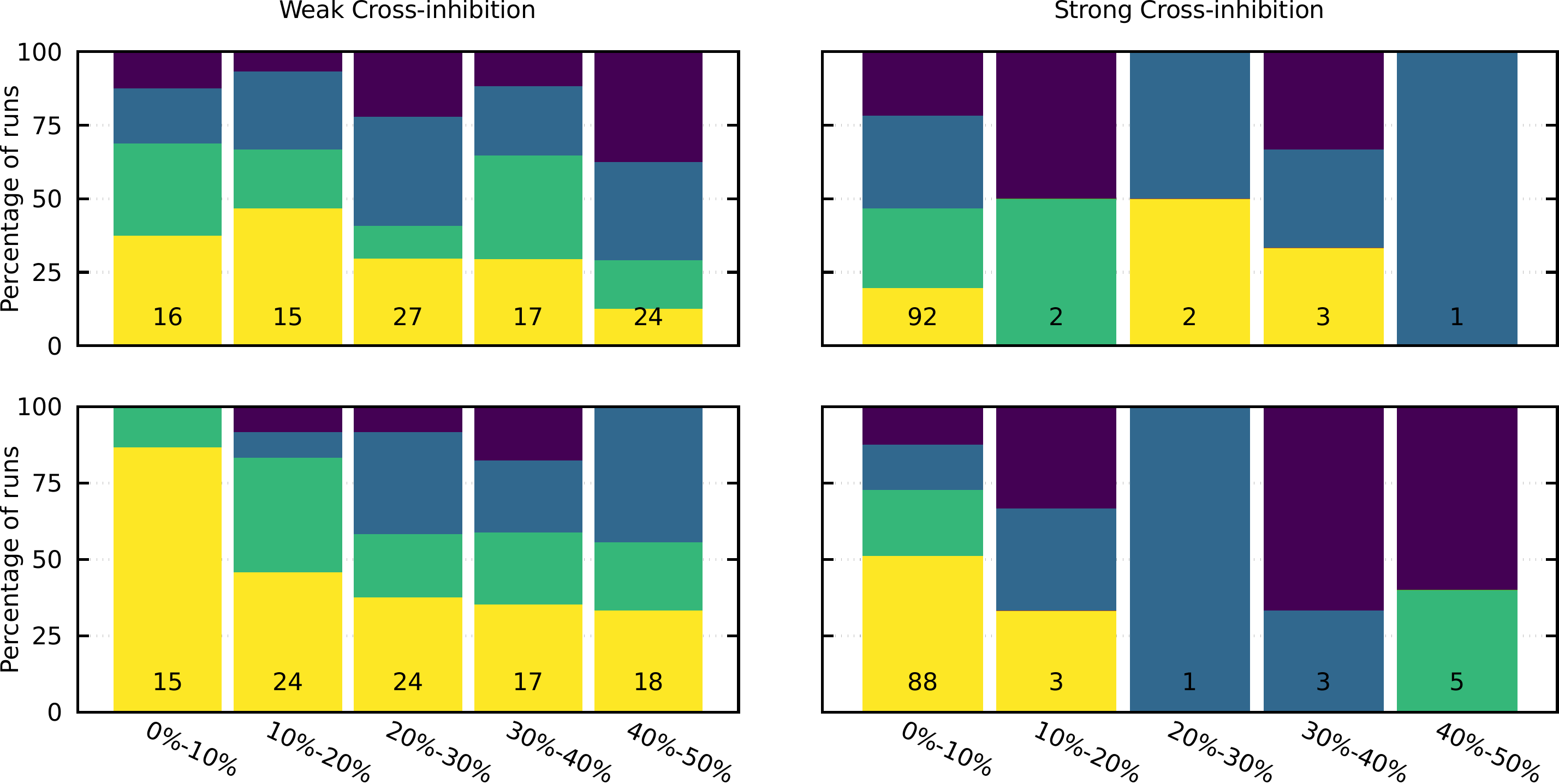}
    \caption{Empirical distribution over 100 runs of the occurrence of the last two words in the vocabulary (see Figure~\ref{fig:cv}) detailed for different distribution of the foraging swarm across the two sources, computed at the time of vocabulary convergence with $P_s = 0.001$. Each stacked histogram corresponds to a specific distribution of robots over the non-selected source ($\frac{\mathcal{P}_X}{\mathcal{P}_O+\mathcal{P}_X}$). Bars are colour-coded as in Figure~\ref{fig:cv}. Over each histogram, the number of runs that resulted in the specified range is displayed. All tested values of $P_s$ present similar results, shown in Figure~S2.
    In the rare case of an equally split swarm ($\mathcal{P}_O=\mathcal{P}_X $), there is no notion of matching an non-matching words. In that case, we redistribute $\AAn$ and $\BBn$ equally between $\OO$ and $\XX$ (one half each). Similarly, $\ABn$ and $\BAn$ are redistributed equally to $\OX$ and $\XO$.  Top row: classic game. Bottom row: spatial game.}
    \label{fig:cvSpread}
\end{figure}

To better understand the relationship between source selection and
naming dynamics, in Figure~\ref{fig:cvSpread} we show how the
distribution of agents between sources relates with the provenance
of the last two words in the swarm vocabulary. Indeed, there is a
large difference between a swarm that forages from a single source
and one that instead is evenly split between the two sources.
In the former, we expect vocabulary matching, that is, only words from
the selected source are retained (hence, case $\OO$ and to some
extent $\OX$).
In the latter, we instead expect vocabulary completeness, that is,
words coming from both sources are present (hence, cases $\OX$ and
$\XO$) because both sources are still exploited by the swarm and the
selected source can change over time.  
Indeed, the swarm does not clearly favor the exploitation of any
source, to the point of possibly changing its selected source
overtime, and multiple times.\footnote{Recall that the distribution of
  robots can change over time, and always converges to the selection
  of one source, although after a very long time as discussed in
  Section~\ref{sec:indiv-coll-behav}. Here, we consider the
  distribution at the time of convergence of the naming dynamics,
  which is determined by the probability of speaking $P_s$. Hence, an
  even distribution of robots among the sources is observable not
  only with weak cross-inhibition ($P_\rho=0.1$, see
  Figure~\ref{fig:nss_end}), but also for strong cross-inhibition when
  high values of $P_s$ cause a quick convergence of the vocabulary.}

When the classic game is played, the distribution of robots across
sources has little to no impact on the provenance of the last two
words (top row of Figure~\ref{fig:cvSpread}, see also the
supplementary Figure~S2 for other values of $P_s$).
For the spatial game, instead, vocabulary matching is observed when
the swarm has clearly selected one of the sources. Conversely,
vocabulary completeness is more often observed with swarms that are
still exploiting two sources. This is evident in case of weak
cross-inhibition that entails slower dynamics in the source
selection process. With strong cross-inhibition, the swarm quickly
converges to exploiting a single source, and the cases in which the
swarm is exploiting both sources at the time of convergence are very
rare. Only when the language dynamics are particularly fast we can
observe cases of vocabulary completion for strong cross-inhibition, as
shown in supplementary Figure~S2 for $P_s=0.002$.

From this analysis we can conclude that the spatial game leads to 
language dynamics that correctly represent the sources
relevant to the swarm, that is, those from which the swarm is
currently foraging. This is obtained solely by the creation of words,
which is strongly correlated with the source discovery. The interplay
between language and foraging dynamics preserves such correlation
despite the high number of interactions between robots from different
populations and with different vocabularies. In the next section,
we study how this is possible by looking at the interaction patterns
between robots.

\section{A study of the swarm's spatial characteristics}
\label{sec:study-swarms-spatial}

There are two extremes for the swarm to reach convergence on a final
word. Either the swarm converges as a whole---homogeneously---on this
final word, or sub-populations foraging from different sources first
converge toward a word representing their source, and then a
competition between these two words determines the final outcome.
In this section, we look
at how robots create and share their words, and how they exchange
words within and across foraging sub-populations.

\begin{figure} [!t]
    \centering
    \includegraphics[width=\textwidth]{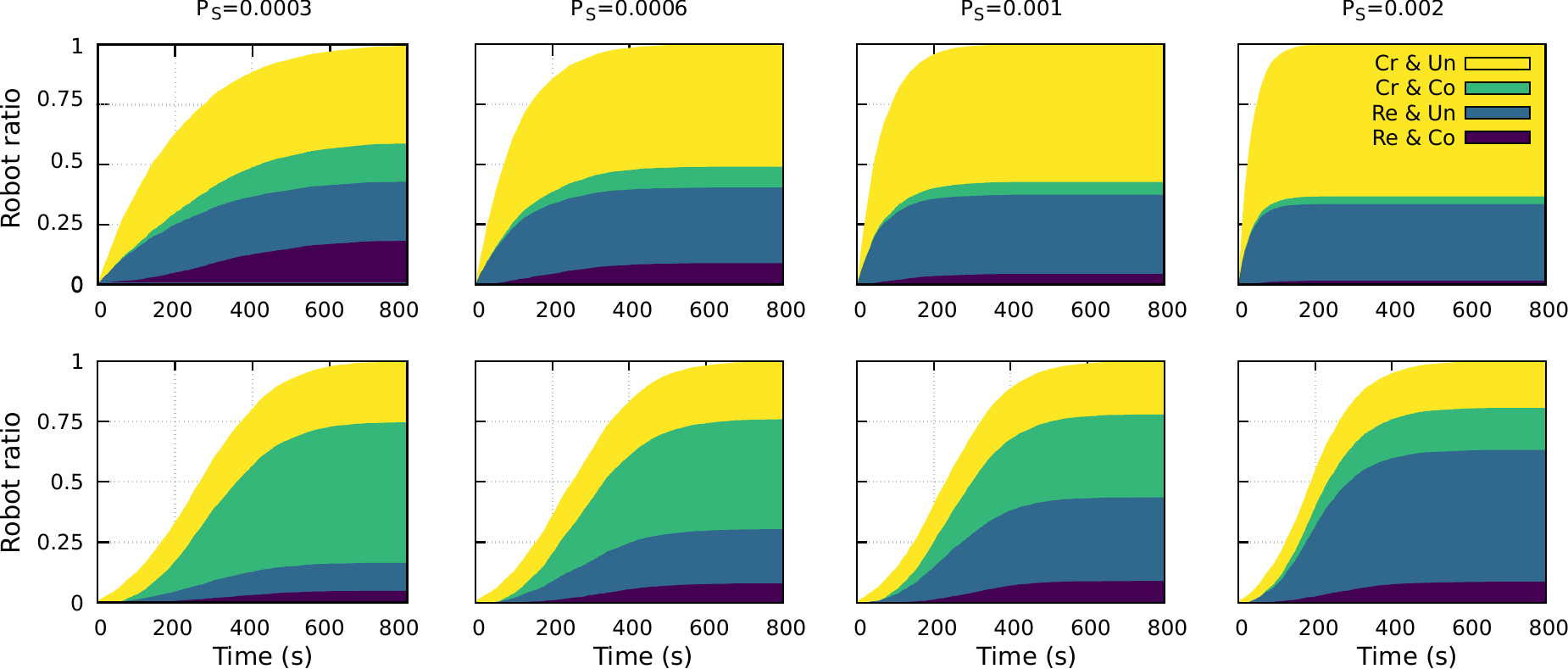}
    \caption{Evolution over time of the origin of each robot's first
      word (weak cross-inhibition). The value of the y
      axis correspond to the ratio of robots having a word in their
      vocabulary. This word can be either created independently by a
      robot (Cr) or received from another robot (Re); and either while
      the robot is uncommitted (Un) or committed (Co). Similar
      dynamics are displayed in the case of strong cross-inhibition
      (see Figure~S3 in supplementary material).  Top row: classic game. Bottom row: spatial game.}
    \label{fig:disco}
\end{figure}

\subsection{Impact of spatial word creation}
First of all, we look at the initial phases of the naming game, when
robots create and share new words.  Indeed, the difference between the
classic and the spatial game is solely related to this phase. 
Besides word creation, robots can fill their vocabulary with words shared by
others. To better understand how robots obtain their first word, we
plot in Figure~\ref{fig:disco} the cumulative number of robots with at
least one word in their vocabulary for the case of weak
cross-inhibition.\footnote{Results for strong cross-inhibition are
very similar and are displayed in Figure~S3.} We highlight 
whether the first word was created by the robot itself or received 
from other robots upon playing the naming game. Finally, we distinguish 
between robots being uncommitted and exploring, or robots committed and 
exploiting one source. Uncommitted robots are particularly relevant, 
as they can get committed to any source, despite having a word associated with
one or the other: they do carry a naming information that may not
correspond to the source they will become committed to.

For the classic game (top row in Figure~\ref{fig:disco}), we note that
the word creation dynamics is rather fast and solely depends on the
probability of speaking $P_s$. Additionally, uncommitted robots
represent the large majority, meaning that word creation is
strongly uncorrelated from source selection: even if a word is
created closer to a source, it is generally associated to an
uncommitted robot that may eventually get committed to any source,
due to recruitment or discovery.\footnote{Recall that robots
periodically return to the home location, where they can get
recruited by any other robot, or they can start a new exploration
trip in a totally different direction from the previous one. Hence,
an uncommitted robot that creates a word near one source may
get recruited to the other source or discover it in the following exploration trip.}

In the spatial game instead (see bottom row in
Figure~\ref{fig:disco}), the dynamics of word creation is independent
of $P_s$ because it is determined by robots encountering a
source. Specifically, $P_s$ does not impact the number of robots that create a word when uncommitted, as these robots individually discover a source following the
foraging dynamics. However, $P_s$ determines the share of robots that
create a word when committed or that receive a word when
uncommitted. The former is higher when $P_s$ is small, as the foraging
dynamics are faster than the language game dynamics, meaning that
several robots get recruited first and encounter a source while still having an empty
vocabulary. These robots have a naming information that is strongly
correlated with the source they are exploiting. Conversely, with
high $P_s$ the number of uncommitted robots that receive a word from
other robots grows. These robots potentially have a naming information
that differs from the source they will exploit, leading to lower
spatial correlation. As a matter of fact, matching and completeness
are slightly worse for this case, as can be observed in supplementary
Figures~S1 and S2.

 \subsection{Communication topology and interactions within the swarm}

\begin{figure}[!t]
    \centering
    \begin{subfigure}[!hb]{\textwidth}
        \includegraphics[width=\textwidth]{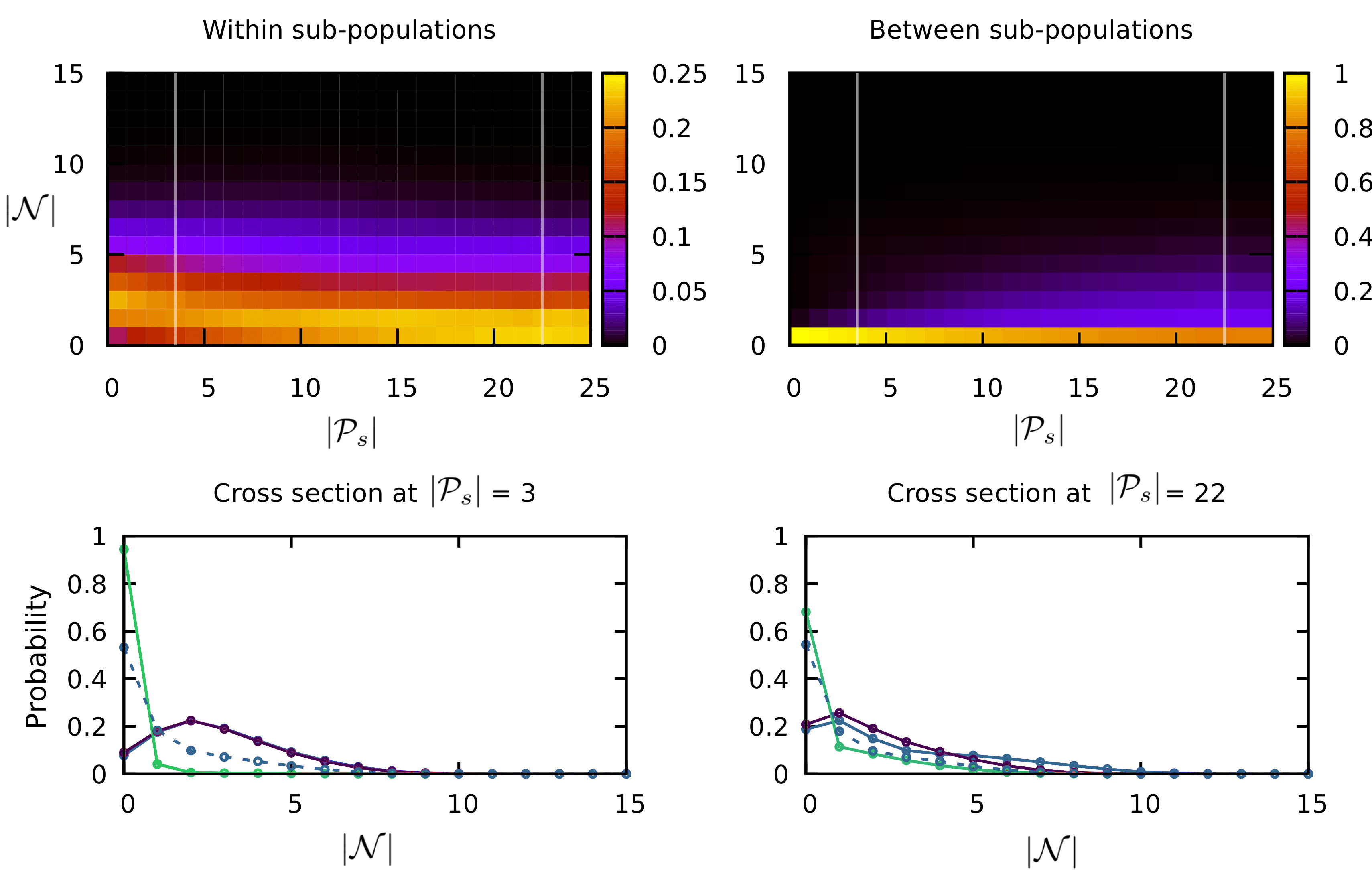}
    \end{subfigure}
    \caption{Top row: the heatmaps represent the probability
      distribution $P_\Sigma$ of each robot's neighbourhood's size
      ($|\mathcal{N}|$, y axis) for each possible partition in
      sub-populations ($|\mathcal{P}_s|$, x axis), limited to
      interactions occurring within a sub-population (top left) or
      between sub-populations (top right). Vertical lines indicate the
      cross-sections displayed in the bottom panels. Bottom row:
      probability of occurrence of each robot's neighbourhood's size
      for $|\mathcal{P}_s|=3$ (bottom left) and $|\mathcal{P}_s|=22$
      (bottom right). The plots represent the probability $P_\Sigma$ of observing
      a neighbourhood size considering interactions within the whole
      swarm (blue), within sub-populations (purple), and between
      sub-populations (green). The dotted-blue line represents the
      case of the whole swarm forced to remain in the exploring state
      (sub-population $\mathcal{P}_U$).}
    \label{fig:neigh}
\end{figure}

Once words have been generated, the MNG imposes a selection process
until a single one is selected. This process takes place through
speaker-hearer interactions, and can be strongly influenced by the
communication topology~\citep{baronchelli2006topology,Moretti:2013wn}.
The latter is determined by the distribution of robots in space, which
is a result of the foraging task the robots carry out. To understand
how the different sub-populations of the swarm interact, we performed
an experiment with locked-size populations, forcing all robots in a
pre-defined committed state. We measure the size of the neighbourhood
$\mathcal{N}$ with which robots can potentially interact anytime, and
we further distinguish between neighbours belonging to the same or to
a different population. In Figure~\ref{fig:neigh}, the probability of
observing a neighbourhood of a given size is displayed for each
possible partition $\mathcal{P}_s$ between sub-populations, where
$\mathcal{P}_s= p$ indicates that $|\mathcal{P}_A|=p$ and
$|\mathcal{P}_B|=N-p$ (in these tests, $\mathcal{P}_U=\emptyset$). Additionally,
we also consider the case in which $|\mathcal{P}_U|=N$, where robots
are forced in the random exploration state. Given a population of agents $\mathcal{P}$, the probability of observing a given neighbourhood of size $n$ exclusively composed of agents from population $\mathcal{P}'$ is computed as follows:
\begin{equation}
  P_\Sigma(|\mathcal{N}| = n | \mathcal{P}) = \frac{1}{T |\mathcal{P}|} \sum_t \sum_{r\in \mathcal{P}'} H_r(n,t,\mathcal{P}'),
\end{equation}
where $H_r(n,t,\mathcal{P}')$ counts the timesteps $t\leq T$ in which
robot $r$ has neighbourhood of size $n$ limited to robots belonging to
population $\mathcal{P}'$. Hence, this probability is strictly
dependent on the size of the populations $\mathcal{P}$ and $\mathcal{P}'$ that are being
considered.

For small values of $|\mathcal{P}_s|$, one of the sub-populations is
large and interactions within sub-population dominate (see
Figure~\ref{fig:neigh}, left panels). The neighbourhood size can take
large values (e.g., more than 5 robots), even larger than the case of
randomly exploring robots (see Figure~\ref{fig:neigh}, bottom-left panel).
Contrarily, interactions between sub-populations are practically
absent, the typical neighbourhood size being $|\mathcal{N}| = 0$ (see
Figure~\ref{fig:neigh}, top-right panel). The more the partition among
sub-populations is even, the more frequent the interactions among
sub-populations become. Still, robots more likely interact within the
same population, and only few cross-population interactions are
recorded (see Figure~\ref{fig:neigh}, bottom-right panel).
This confirms that, if the swarm leans towards selecting a single
source, the language dynamics are played mostly within the same
population, reinforcing the correlation between words and sources in
favour of matching. At the same time, the small number of interactions
between sub-populations also favour completeness, with each
sub-population having the chance to converge on its own word.

\begin{figure}[!ht]
    \includegraphics[width=\textwidth]{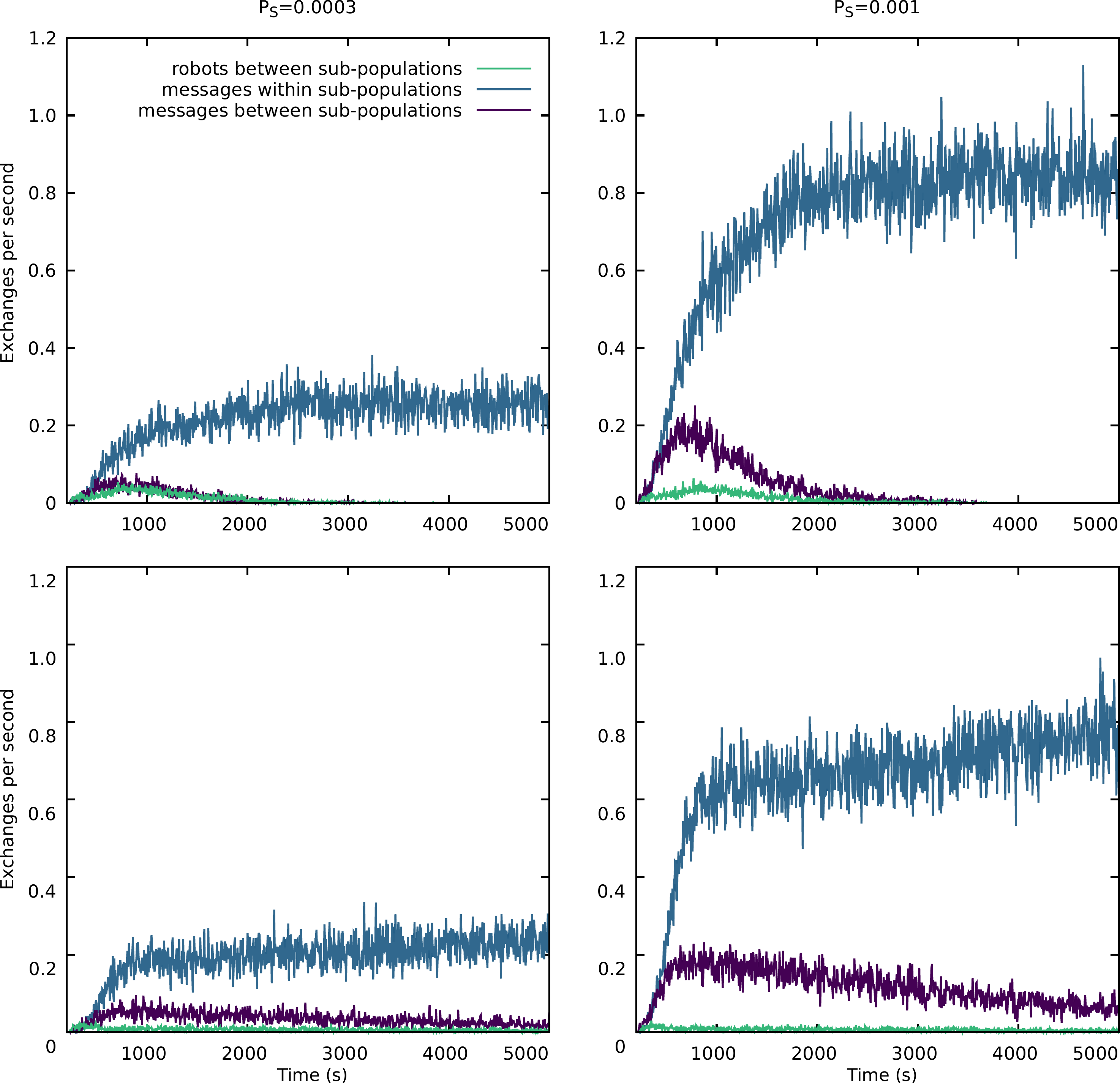}
    \caption{Evolution over time of the rate of communications within
      and between sub-populations exploiting different sources, and of the rate of robot movements
      between sub-populations. Each graph has been plotted for the
      spatial game.  Similar dynamics are displayed by the classic
      game (see supplementary Figure~S4).  Top row: strong cross-inhibition. Bottom row: weak cross-inhibition.}
    \label{fig:xChange}
\end{figure}

It is worth recalling that, besides communications between
sub-populations, a mismatching word can enter a sub-population also when
it is physically carried by a robot changing from one to the other
population. In order to understand how relevant the movements of
robots between sub-populations are for the spreading of words, we
measured the rate at which these movements take place, and compared it
with the rates of interactions within and between populations during a
standard experiment (see Figure~\ref{fig:xChange}). The results
indicate that movements between sub-populations are not as
frequent as the interactions via message exchange, especially when the
probability of speaking $P_s$ is high (see also Figure~S4). Indeed,
the rate at which messages are exchanged within and between
populations increases with $P_s$, and is generally larger for
intra-population interactions, confirming our previous
analysis. Conversely, the rate at which robots move from one
population to the other does not depend on $P_s$, and is higher when
cross-inhibition is strong. We infer that the movements of robots 
between sub-populations do not have a relevant impact on the 
language dynamics in this specific experimental setup.

In the light of the presented results, we can conclude that the
pattern of interactions between robots favours the segregation between
sub-populations. This means that different words are likely selected
within each sub-population, resulting in the vocabulary
completeness. At the same time, vocabulary matching is possible thanks
to the strong correlation between word creation and source
exploitation by committed robots, as discussed above. While the vocabularies well
represent the environmental features and their relevance for the
swarm, we note that completeness is a transient property.
Indeed, the MNG dynamics determine the convergence towards a single
word shared by the swarm, loosing information about previously
exploited sources. To avoid this, we present in the next section a
proof of concept of a language game to preserve matching and complete
vocabularies.

\section{Emergence of spatial categories for foraging swarms}
\label{sec:from-vocab-dict}

Keeping a complete description
of the environment with all its sources requires the ability to
distinguish between different regions in space, leading to the
construction of spatial categories. 
We consider a spatial category as a set of possible words, associated to an area representing the region covered by the category (here, a circle defined by its radius and its center, the latter determining the prototype location of the spatial category).
Speaking in general terms, any location in space can belong to one category, to multiple ones (in case of
overlapping categories) or to none (in the case of a non exhaustive
partition of the space). The same robot can potentially hold multiple
words (synonyms) referring to a given category. As a consequence, the set of
the categories known to a robot---and, by extension, to the swarm---results in a kind of thesaurus. In
this section, we propose a language game based on word-location pairs
with the goal of representing the landscape of available
sources. The language game is now first played on categories and
then on words, making it more similar to a category game \citep{baronchelli2010modeling}.


\subsection{Experimental Setup} 
Similarly to the spatial game discussed above, categories---and associated names---are spontaneously created when a robot encounters a source at a location that is not represented by any available category. Even if a category exists for the same source, a robot may enter from a location that
is not covered by the current category description. This leads to an
initial proliferation of categories, which are subsequently pruned by
a merging mechanism (see below).
 
With probability $P_s$, a robot knowing at least one category takes the speaker role: it first
selects one of its known categories, followed by a word belonging to this category's inventory.  The speaker will share with the neighbours the
selected word paired with the category prototype's location. In order to
maintain a correspondence between the foraging behaviour and the
language game, the selection of the category is determined by the
commitment status: the speaker always selects the category
corresponding to the sources it is foraging from. For uncommitted robots, the category is selected randomly. On the hearer side, first a match of the received word-location pair
must be found with the known categories.  If the location does not
belong to any known category, the hearer creates a category centered
on that location, with a default starting radius of
$r_{0} \in \{0.2, 0.3, 0.4\}$, and add the received word to this
category. If the location belongs to only one category, the MNG is played as previously described (see
Section~\ref{sec:indiv-coll-behav}) with respect to the matching
category's inventory. If the word is fitting multiple categories,
these are merged into one, and then the MNG is played
with respect to the resulting category's inventory.  Categories are
merged two by two, with the resulting category being the smallest
possible circle containing each original category's circle. The merged
vocabulary is the union of each category's vocabulary.



To evaluate the ability of the swarm to generate shared spatial
categories that correctly represent the available source landscape,
we performed a series of experiments varying both the probability of
speaking $P_s$ and the value of the initial category radius $r_0$.  We
introduce no change in the physical layout of the arena (see
Figure~\ref{fig:arena}).  Experiments are run for longer times, and
are stopped once convergence is reached on both categories and number of words in each category. The additional complexity introduced by categories entails a slower language dynamics with respect to the simple naming game described before.
%
%
To study the ability of the foraging swarm to correctly represent both
sources, we prevent the selection of a single one by forcing
$P_\sigma=0$. In this way, the robots will find and exploit both
sources (possibly with an uneven distribution across the two), and
no robot will ever change source. As we observed in
Section~\ref{sec:study-swarms-spatial}, the effects on the language
dynamics of robots physically moving from one to the other source
are anyway negligible.

\subsection{Results}

\begin{figure}[!t]
    \centering
    \includegraphics[width=0.8\textwidth]{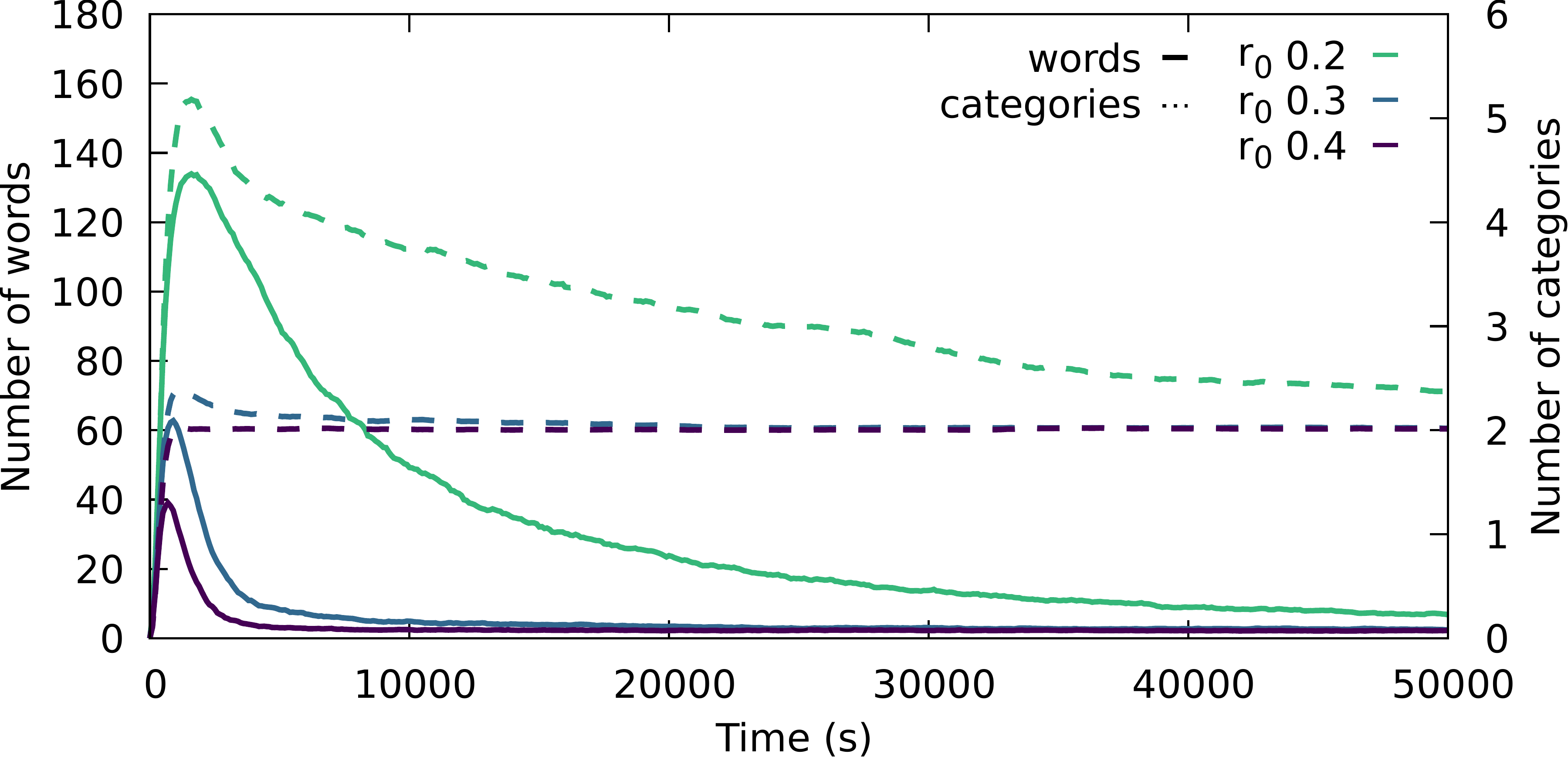}
    \caption{ Average number of different words (solid lines) and different categories (dotted lines) present within the swarm. The dynamics over time are plotted for different values of $r_0$, and for a fixed probability of speaking $P_s = 0.001$. Plots for other values of $P_s$ are available in the supplementary Figure~S6.}
    \label{fig:catEvo}
\end{figure}

The evolution over time of the number of words and of categories is
shown in Figure~\ref{fig:catEvo} for $P_s=0.001$ (see Figure~S6 for
other values). Both words and categories follow a similar pattern,
with an initial fast proliferation and a following convergence toward
the minimum number of elements: one single category for each source,
and one single word per category. The radius $r_0$ determines the
likelihood that a new category is created: when the radius is large
enough, the initial category easily covers the whole source, and
creation of new categories for the same source is unlikely. As a
consequence, also the number of words generated is lower, because
different words are generated for different categories, and the
vocabularies are preserved by the category merging. In any case, the
system tends to converge to the minimum number of words/categories for
each value of $r_0$.
We note that the actual convergence on two categories (and hence two
words) is not always permanent, as new categories can emerge after
convergence on two categories. These rare events are unlikely to have
long lasting impact as the swarm can recover quickly. Under these
conditions, we define as time of convergence (over two categories or
two words) the first time the whole swarm reaches the minimum number 
of words/categories.

\begin{figure}[!ht]
    \includegraphics[width=\textwidth]{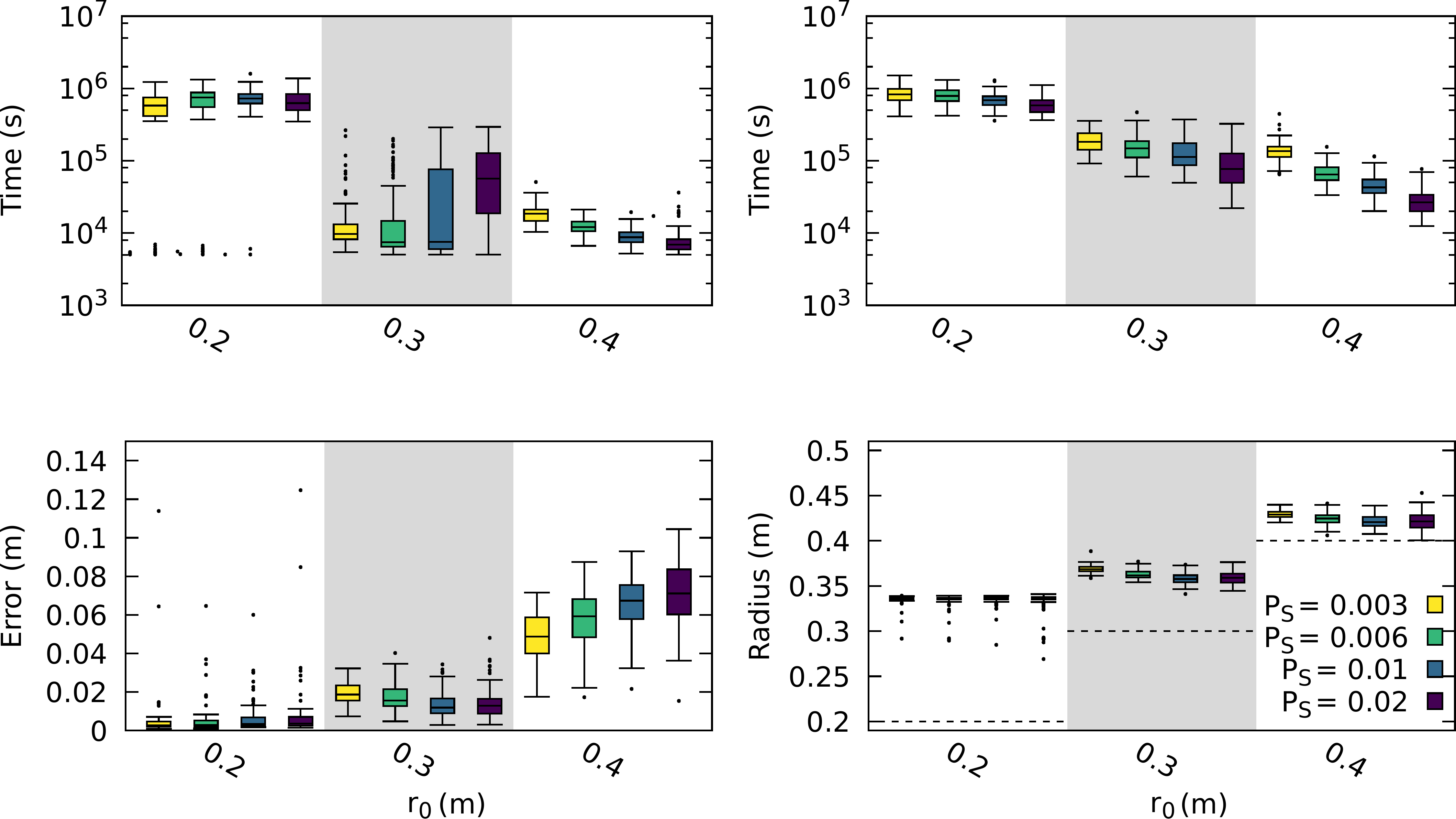}
    \caption{Effects of the initial category radius $r_0$ and of the
      probability of speaking $P_s$. Top left: categories' convergence time. 
      Top right: words' convergence time. Bottom left: average error of
      the final category prototype with respect to the center of the
      associated source. Bottom right: average final radius of each
      category compared with the initial value of $r_0$ (dotted
      line).}
    \label{fig:catBox}
\end{figure}

Both category and word convergence times depends heavily on $r_0$, but also on 
the probability of speaking $P_s$ (see the top-left panel in
Figure~\ref{fig:catBox}). When $r_0$ is intermediate-small, the large
proliferation of categories requires several merging operations, and
having more variability in each category does not give an
advantage. On the other hand, for large $r_0$ few categories are
formed, and a high $P_s$ helps in quickly converging. These dynamics
are confirmed also by the time of convergence to a single word per
category (Figure~\ref{fig:catBox} top right), which always decreases 
when $P_s$ increases, with a larger effect for larger $r_0$.

Apart from the speed of convergence, another relevant aspect concerns
the accuracy with which the emerging categories describe the sources
to which they are associated. To measure this, we consider the
position error as the distance between the center of the category and
the center of the source (see Figure~\ref{fig:catBox} bottom left)
and the average radius of the final category (Figure~\ref{fig:catBox}
bottom right). When the initial radius is smaller, the error in the
position of the prototype is very small, as it results from the
average of many categories defined all around the source. With
larger $r_0$, the position error increases because fewer categories are
generated. Large values of the probability of speaking $P_s$ result
in even fewer generated categories: as robots receive their initial category
from other robots, larger errors are made.  
For what concerns the final radius of the emergent categories, smaller
values are observed for small $r_0$. However, the relative increase of the final radius with respect to the initial $r_0$ is much larger for small $r_0$ than for large $r_0$ because many different categories are merged together.

\section{Conclusion}
\label{sec:conclusion}

In this article, we studied how the language game dynamics are
influenced by the evolving topology of a swarm engaged in a
decision-making and foraging task. In particular, we studied how well
the swarm could maintain a description of its whole environment that
is at the same time correct and complete, with the vocabulary
containing only words that are relevant to the swarm, that is, those
associated to sources under exploitation. We focused on such a compelling research question, without questioning properties commonly studied in swarm robotics such as robustness or scalability. Such properties have been largely studied for foraging, language dynamics and decision-making in previous studies and in conditions very close to the ones discussed here \citep{Trianni:2015if,Reina:2015gs}. Hence, they are no further debated, allowing us to focus on the interplay between language and decision dynamics.

We began by comparing two variations of the MNG. One that binds the
creation of words with the sources available in the environment
(spatial game), the other without such spatial correlation (classic
game), where words are used as simple tokens. Note that a mild spatial
correlation is available between words and sources also in the
classic game, given that words are created by robots at locations that
are always closer to one of the sources. However, this was not
sufficient to guarantee the emergence of a correct and complete
language as in the spatial game. Indeed, the stronger correlation
between creation of words and source location granted by the spatial
game is not the only reason for the better matching and completeness.
We observed that a major difference is given by the role of
uncommitted, exploring robots into the creation and sharing of
words. These robots can end up choosing any source, bringing words
created near one source to the population exploiting the other.
Additionally, we observed that the topology of the robot's interaction
network---determined by the robot's movements during the foraging
activity---consists of two almost segregated sub-populations, with
sporadic interactions constrained to the central nest area. Such
segregation creates the conditions for the maintenance of one word for
each source, supporting completeness of the evolving vocabulary.
In order for the swarm to maintain a complete description of the
environment even when sources are not relevant any more, we proposed
as a proof of concept a simple version of a category game embedded in
space. In this setup, the swarm creates different categories for each
source, and ends up retaining an exhaustive description that can also
be sufficiently precise to potentially support the foraging
activities.

One potential drawback of language evolution as observed in our
experiments is related to the time required for emergent conventions
to settle, which can be very large if interactions are sporadic, as
well as the possibility that new conventions enter the population and
destabilize the language dynamics. In this respect, it is important to
note that linguistic conventions do not have an intrinsic value (e.g.,
every name can be equivalent as long as it is understood), but are
more valuable when they are largely shared within a population,
favouring coordination and avoiding misunderstandings. Hence, it is
possible to speed up convergence toward a shared convention within a
population by means of positive feedback mechanisms that favour the
conventions more commonly found within the population. For instance,
the simple rules of the naming game could be enhanced with estimates
of the frequency of words in the population, allowing to favour the
selection of more frequent words when speaking, hence speeding up
convergence.  Additionally, decentralised quorum sensing approaches
can be exploited to determine a final convention, avoiding that
noise is added by new alternatives when a largely shared one is
already present. These and similar mechanisms can reduce the number of interactions required to achieve language convergence within a population, making language games practicable in realistic settings beyond the abstract scenario studied in this paper.

Overall, we believe that merging language dynamics with the
self-organising behaviour of robot swarms can have a high potential,
as the robot behaviour can exploit the emergent descriptions of the
environment in a way that is dependent on the features relevant for
the swarm.  Our experiments demonstrate a possible way to obtain a
meaningful link between tasks and the evolving language, supporting
future research activities. The link between language and behaviour
was relegated here to the creation of words/categories. However,
stronger links can be built if behavioural decisions can be determined
by the evolving language. This also allows to adapt the language to
the environmental contingencies encountered, possibly enabling more
flexibility in the swarm behaviour with respect to changing
environmental conditions \citep{CambierEtAl:2021}.

In future studies, besides describing the relevant features of the
environment, linguistic conventions can be exploited also to agree on
the best course of action for the swarm. For instance, robots would
share short term plans described as a sequence of linguistic elements,
creating and merging them following shared compositional
strategies. In this sense, the possibilities offered by language
evolution are vast, allowing robot swarms to autonomously find
sentence-like solutions to complex tasks made of several
spatially-distributed and temporally-dependent sub-tasks.



\section*{Acknowledgements}
A. Reina acknowledges the financial support from the Belgian F.R.S.-FNRS, of which he is a Postdoctoral Researcher, and from the Office of Naval Research Global (ONRG) through grant no. 12547352. M. Dorigo acknowledges support from the Belgian F.R.S.-FNRS, of which he is a Research Director. V. Trianni acknowledges partial support from the Office of Naval Research Global (ONRG) through grant no. N62909-18-1-2093 and from the European project TAILOR  (H2020-ICT-48 GA: 952215).

\bibliographystyle{abbrvnat}
\bibliography{biblio.bib}
    
\clearpage

\setcounter{figure}{0}
\renewcommand{\thefigure}{S\arabic{figure}}


\begin{figure}[!h]
  \begin{subfigure}{\textwidth}
    \includegraphics[width=\textwidth]{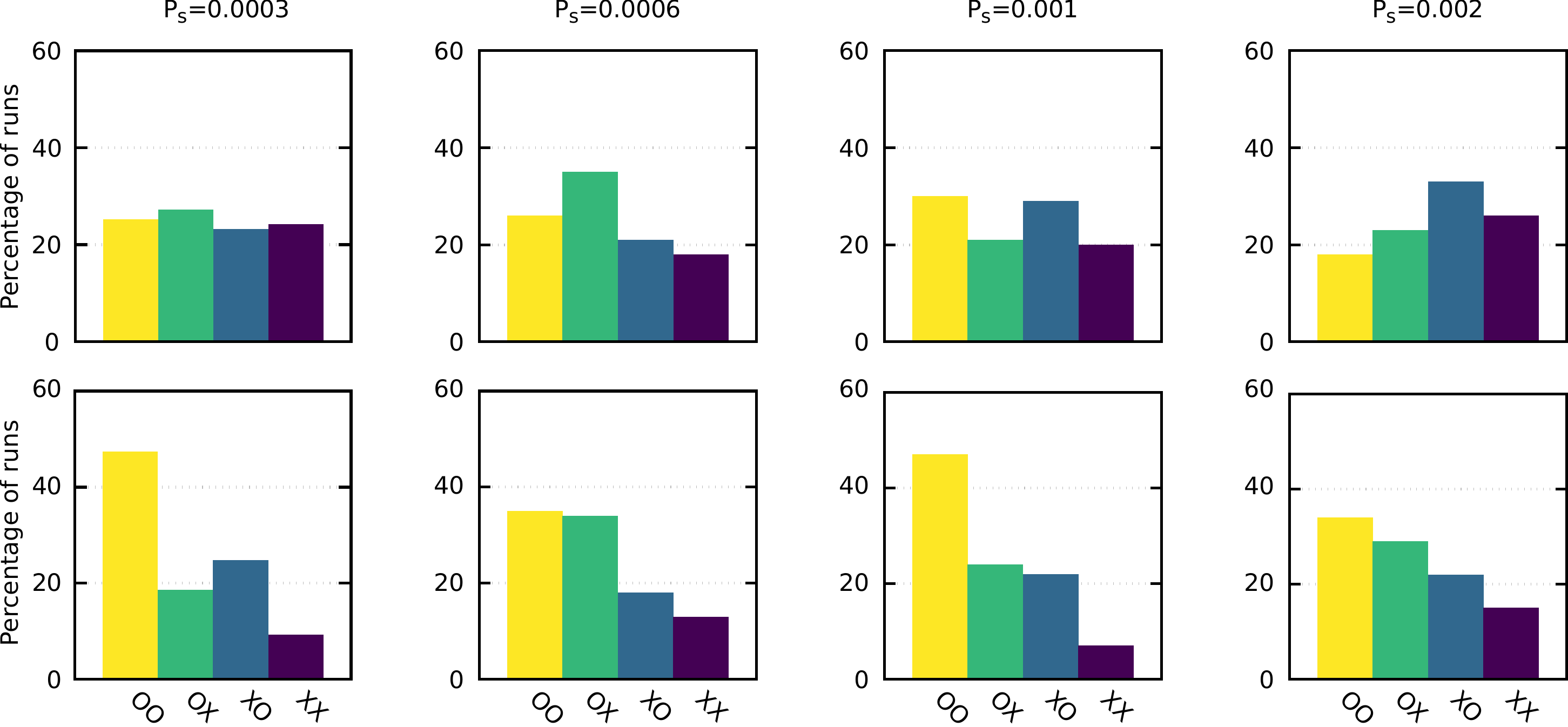}
    \caption{Weak cross-inhibition} \label{fig:suppHisto-a}
  \end{subfigure}%
  \vskip 3em
  
  \begin{subfigure}{\textwidth}
    \includegraphics[width=\textwidth]{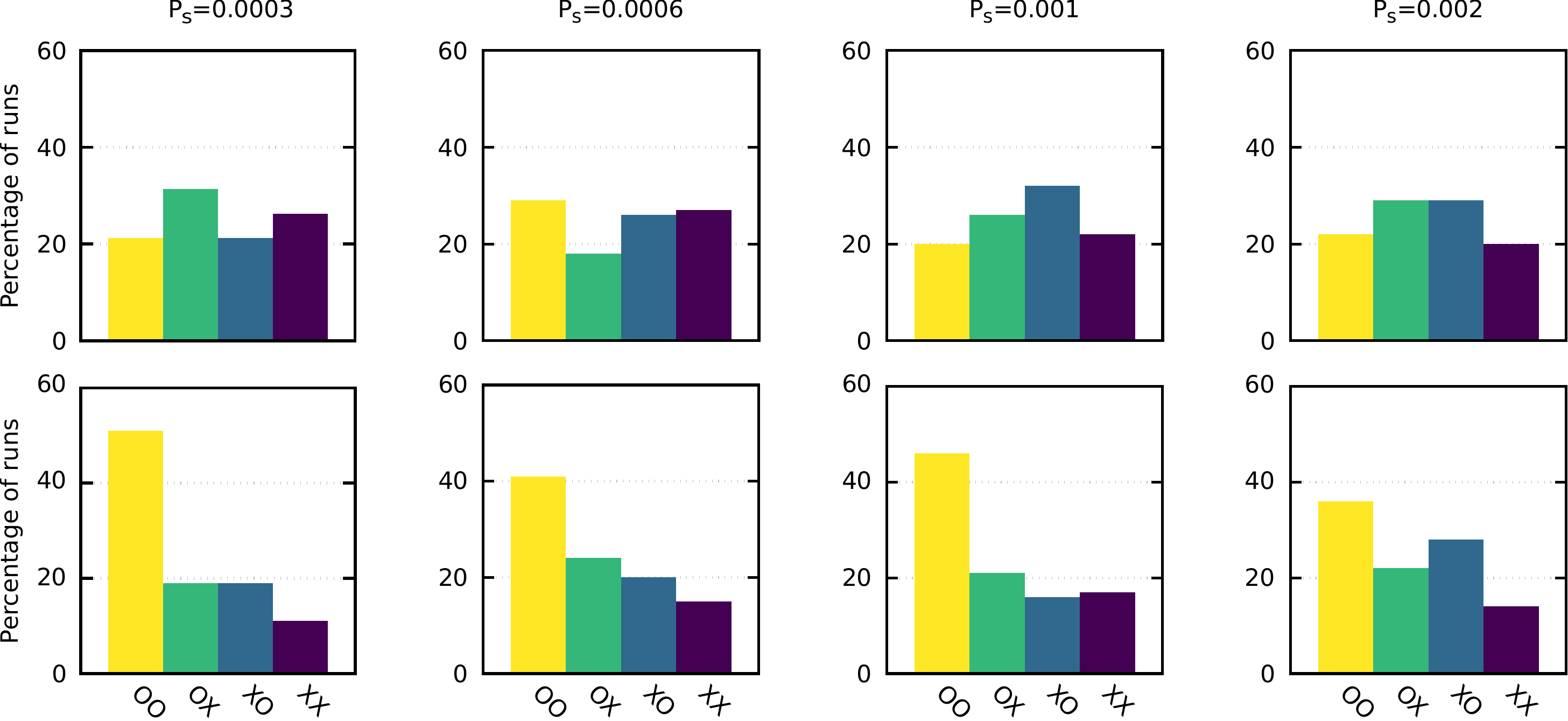}
    \caption{Strong cross-inhibition} \label{fig:suppHisto-b}
  \end{subfigure}%

  \caption{Frequency of occurrence of the last two words in the
    vocabulary within the four identified classes with respect to the
    words matching (O) or not (X) the selected resource ($\OO$,
    $\OX$, $\XO$ and $\XX$). Top row: classic game. Bottom row: spatial game.}
    \label{fig:suppHisto}
\end{figure}


\newpage
\begin{figure}[]
  \begin{subfigure}{\textwidth}
    \includegraphics[width=\textwidth]{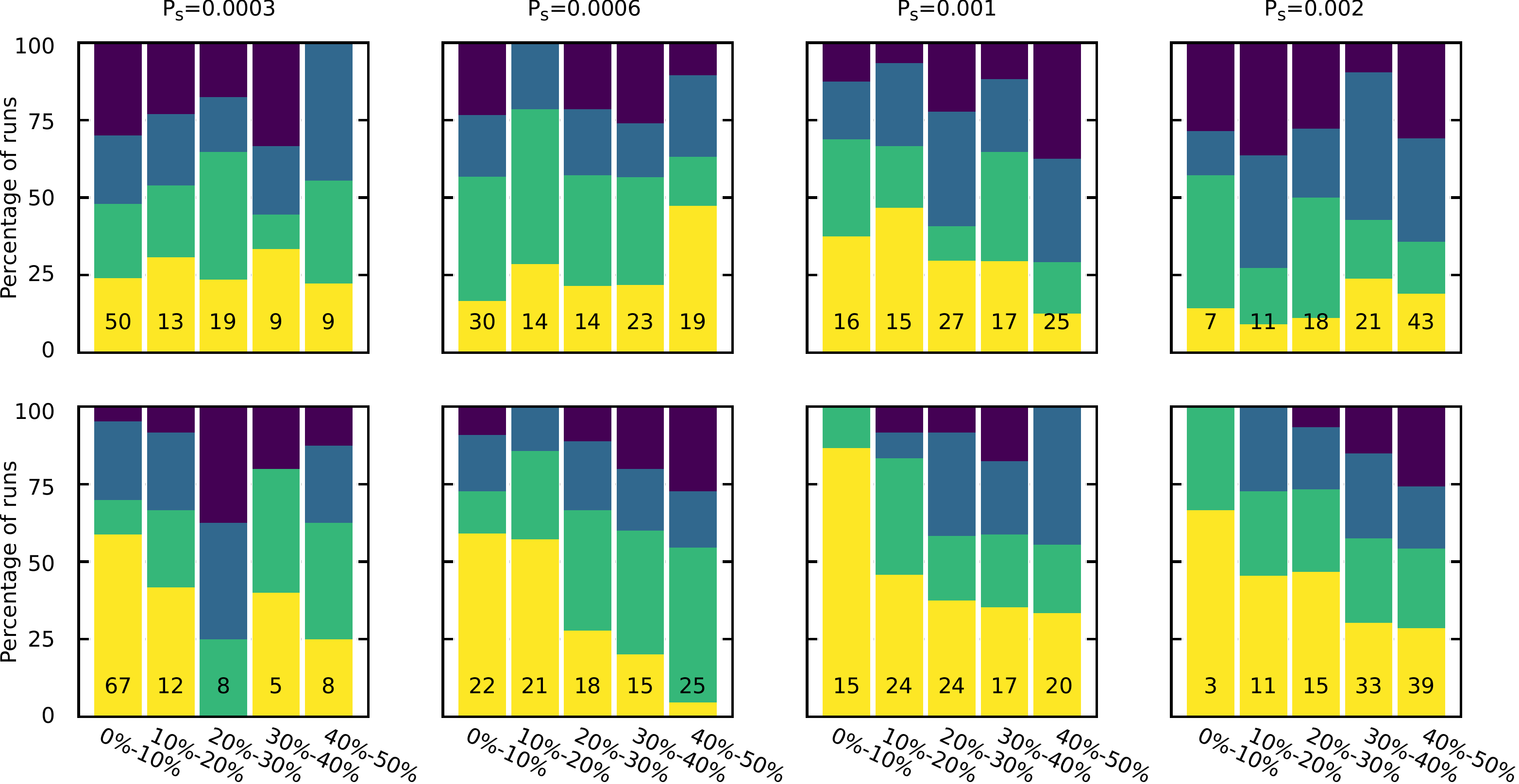}
    \caption{Weak cross-inhibition} \label{fig:suppEndHisto-a}
  \end{subfigure}%
  \vskip 3em
  
  \begin{subfigure}{\textwidth}
    \includegraphics[width=\textwidth]{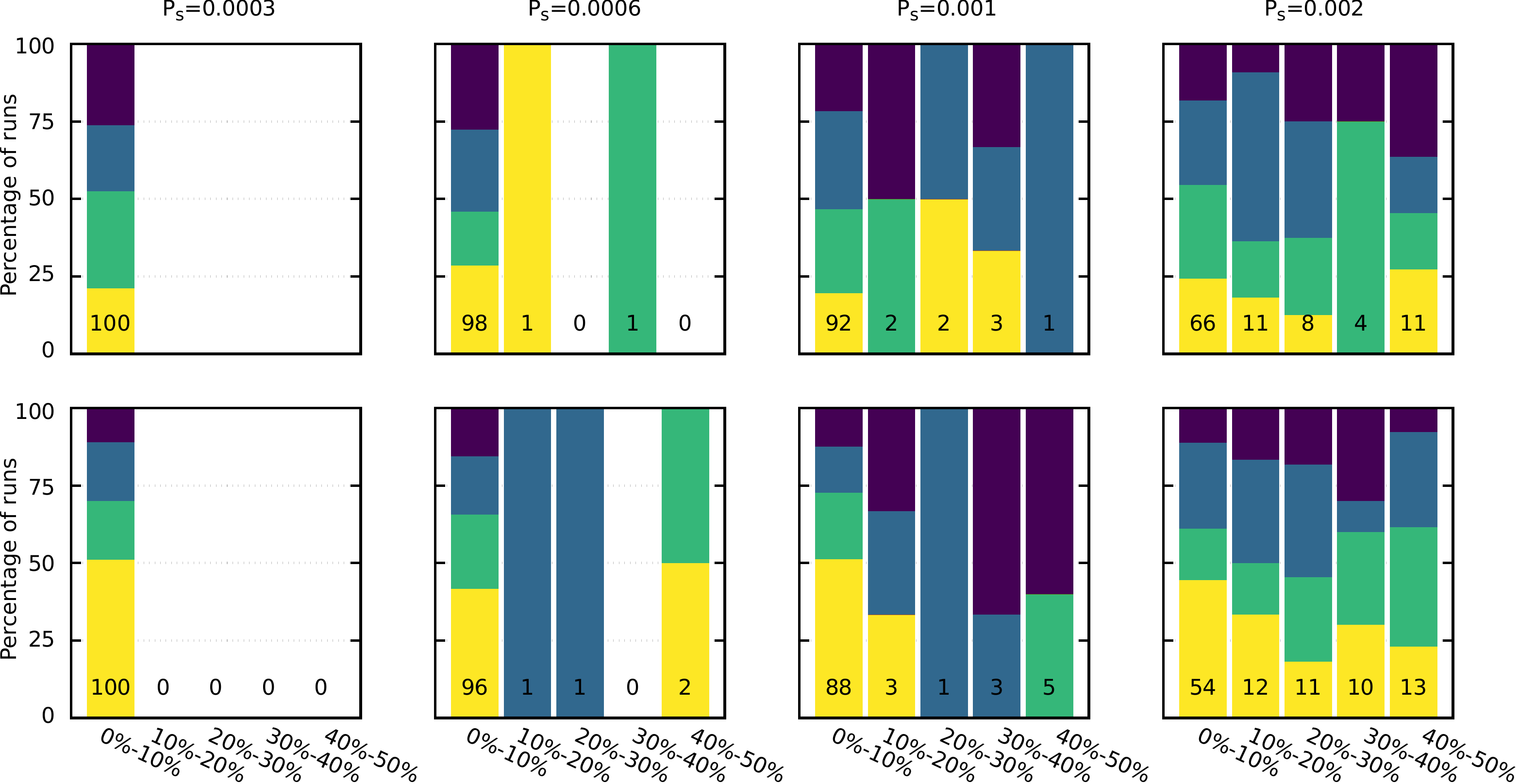}
    \caption{Strong cross-inhibition} \label{fig:suppEndHisto-b}
  \end{subfigure}%

  \caption{Frequency of occurrence of the last two words in the
    vocabulary (see Figure~\ref{fig:suppHisto}) detailed for different
    distribution of the foraging swarm across the two resources,
    computed at the time of vocabulary convergence. In the rare case
    of an equally split swarm ($\mathcal{P}_O=\mathcal{P}_X $), there
    is no notion of matching an non-matching words. In that case, we
    redistribute $\AAn$ and $\BBn$ equally between $\OO$ and $\XX$
    (one half each). Similarly, $\ABn$ and $\BAn$ are redistributed
    equally to $\OX$ and $\XO$. Each stacked histogram corresponds to a
    specific distribution of robots over the non-selected resource
    ($\frac{\mathcal{P}_X}{\mathcal{P}_O+\mathcal{P}_X}$). Bars are
    colour-coded as in Figure~\ref{fig:suppHisto}. Over each histogram, the
    number of runs that resulted in the specified range is
    displayed. Top row: classic game. Bottom row: spatial game.} 
    \label{fig:suppEndHisto}
\end{figure}


\newpage
\begin{figure}[]
  \begin{subfigure}{\textwidth}
    \includegraphics[width=\textwidth]{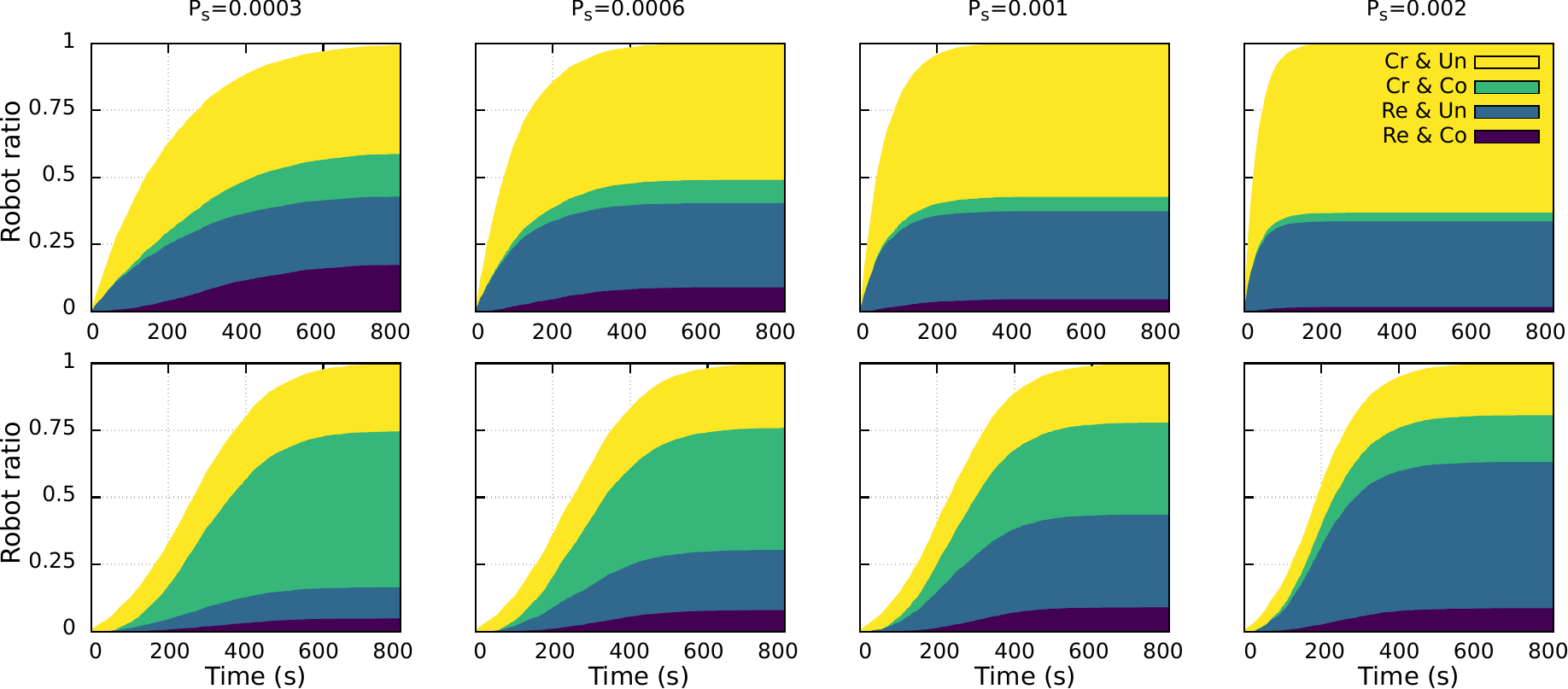}
    \caption{Weak cross-inhibition} \label{fig:cumcreate-a}
  \end{subfigure}%
  \vskip 3em
  
  \begin{subfigure}{\textwidth}
    \includegraphics[width=\textwidth]{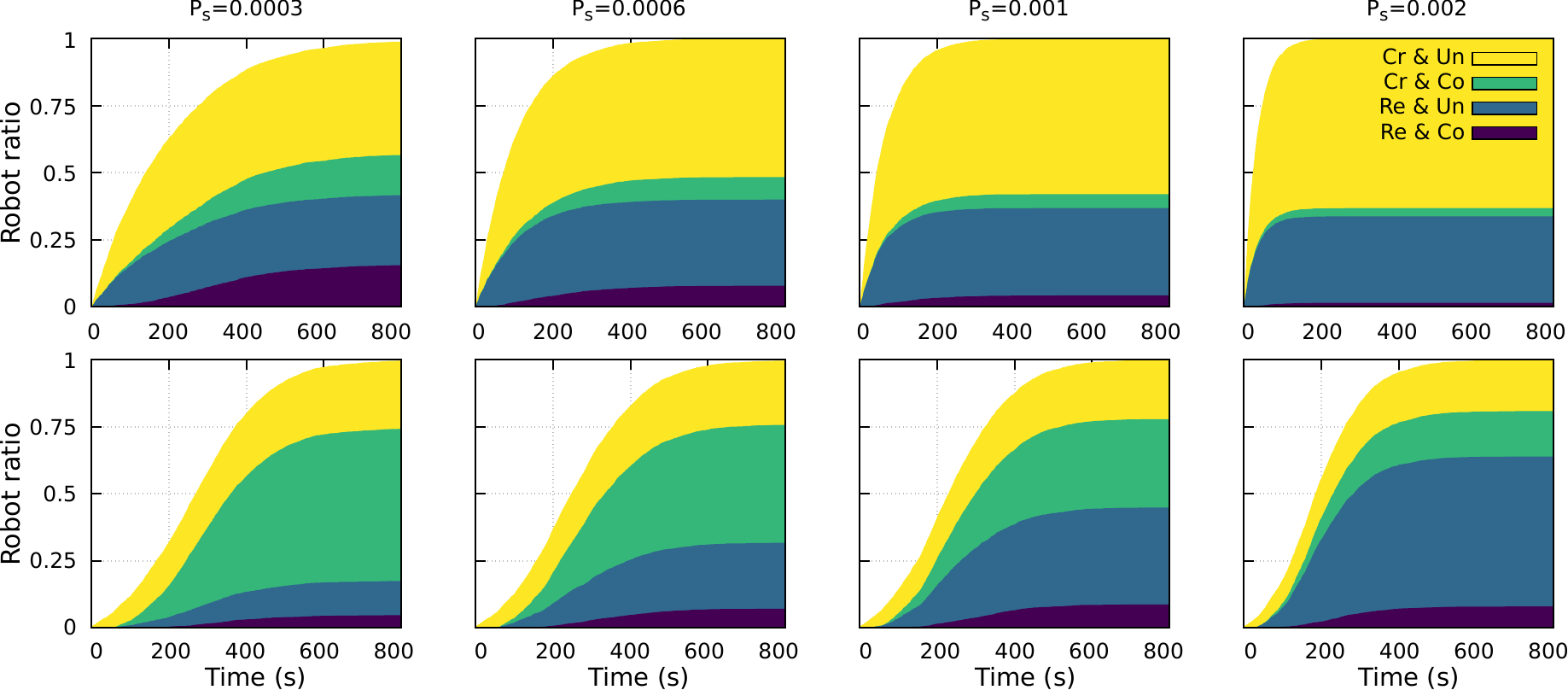}
    \caption{Strong cross-inhibition} \label{fig:cumcreate-b}
  \end{subfigure}%
  
    \caption{Study of the evolution over time of the origin of each
      robot's first word in the case of a weak cross-inhibition. The
      value of the Y axis correspond to the overall ratio of robots
      having a word in their vocabulary. This word can be either
      created upon a discovery (Cr) or received from another robot
      (Re); and either while the robot is uncommitted (Un) or
      committed (Co). Top row: classic game. Bottom row: spatial game.}
    \label{fig:cumcreate}
\end{figure}


\newpage
\begin{figure}[]
  \begin{subfigure}{\textwidth}
    \includegraphics[width=\textwidth]{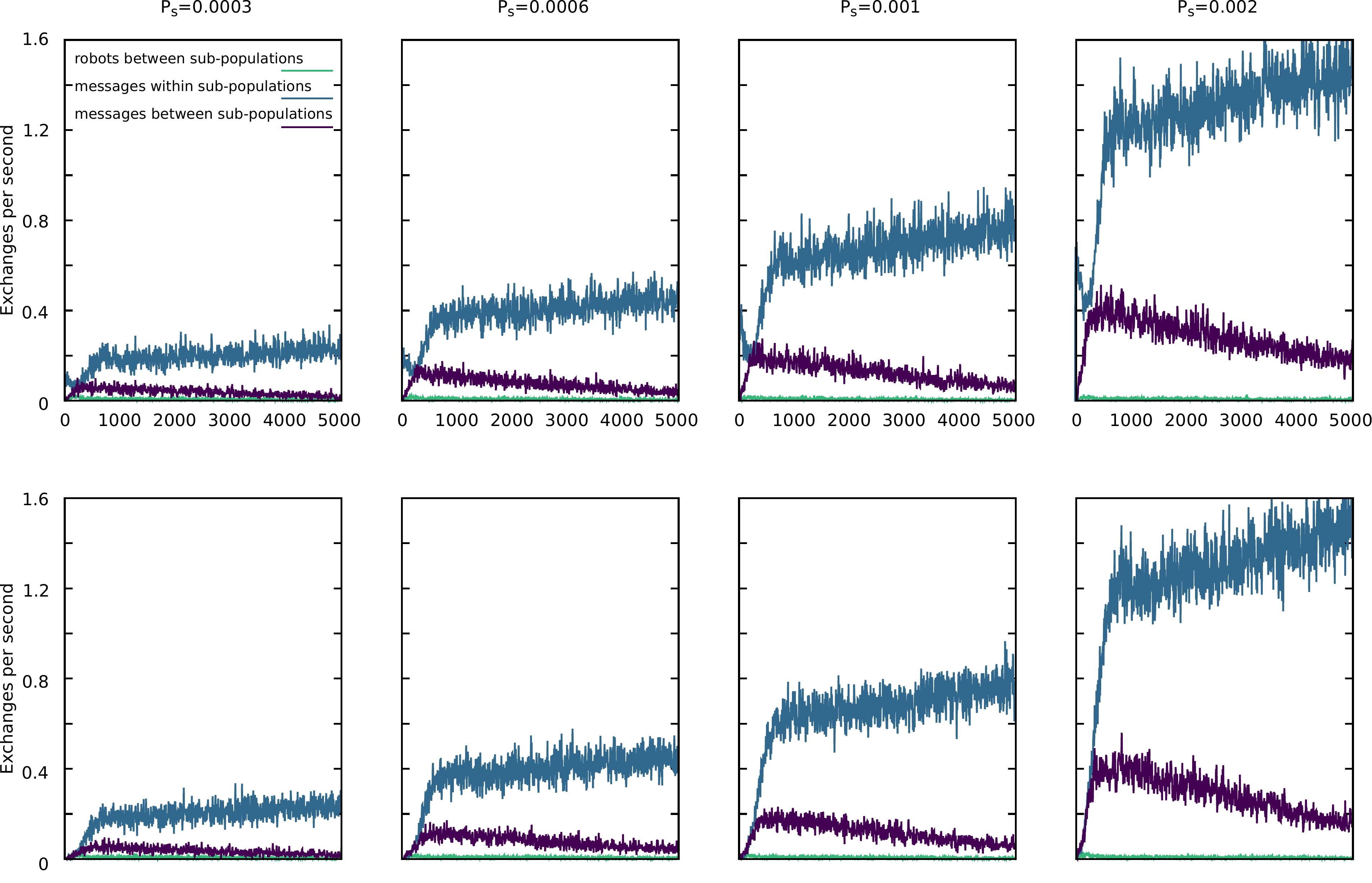}
    \caption{Weak cross-inhibition} \label{fig:commtime-a}
  \end{subfigure}%
  \vskip 3em
  
  \begin{subfigure}{\textwidth}
    \includegraphics[width=\textwidth]{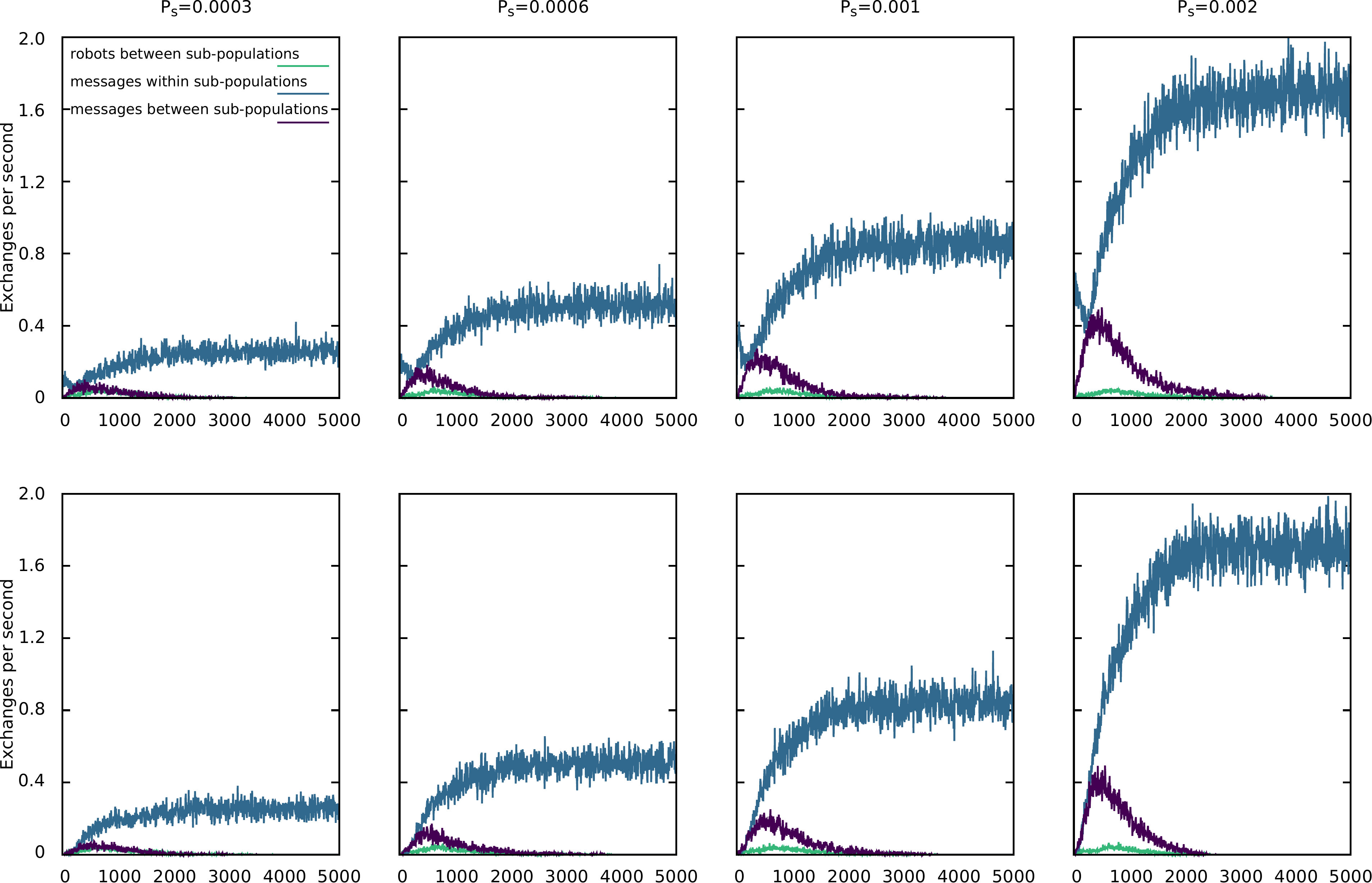}
    \caption{Strong cross-inhibition} \label{fig:commtime-b}
  \end{subfigure}%

  \caption{Evolution over time of the rate of communication within and
    between subpopulations, and of the rate of robot movements between
    subpopulations. Top row: classic game. Bottom row: spatial game.}
    \label{fig:commtime}
\end{figure}


\newpage
\begin{figure}[]
    \includegraphics[width=0.9\textwidth]{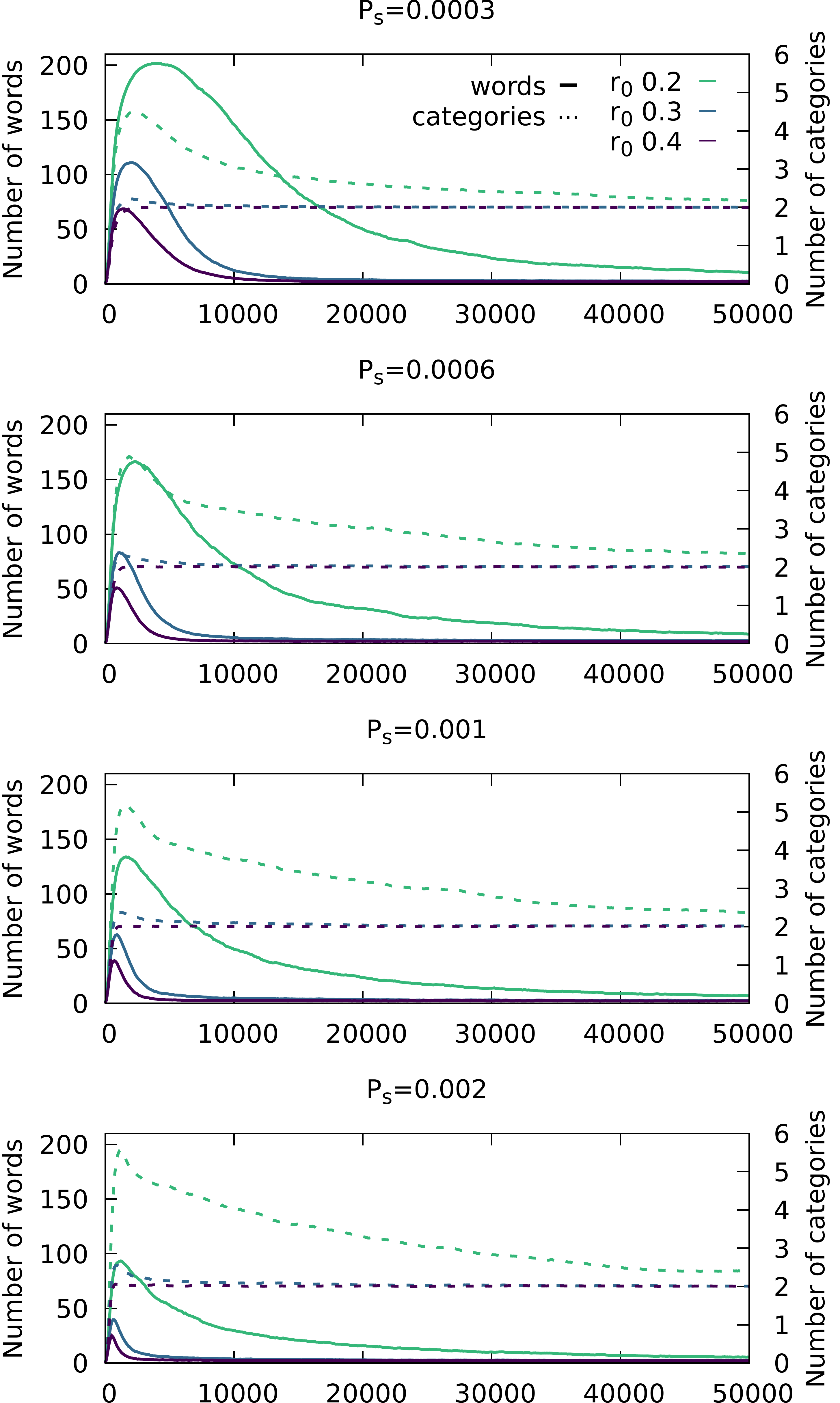}
    \caption{Average number of different words (solid lines) and different categories (dotted lines) present within the swarm. The dynamics over time are plotted for different values of $r_0$.}
    \label{fig:suppCat}
\end{figure}

\end{document}